\documentclass[12pt]{article} 

\usepackage[dvipdfm]{graphicx} 

\usepackage{amssymb}
\usepackage{amsmath}
\usepackage{amscd}
\usepackage{latexsym}
\usepackage{graphicx}

\usepackage{caption}
\usepackage{geometry}
\usepackage{subcaption}

\usepackage{cite}

\newtheorem{definition}{Definition}

\DeclareMathOperator{\re}{Re}

\newcommand{\be}{\begin{equation}}
\newcommand{\ee}{\end{equation}}
\newcommand{\Dlt}{\Delta}
\newcommand{\dlt}{\delta}

\newcommand{\bt}{\beta}

\newcommand{\al}{\alpha}

\newcommand{\gm}{\gamma}

\newcommand{\lbd}{\lambda}

\newcommand{\rgl}{\rangle}
\newcommand{\lgl}{\langle}

\begin{document}

\begin{center}

{\Large{\bf Calibration of Quantum Decision Theory: \\
Aversion to Large Losses and Predictability of Probabilistic Choices} \\ [5mm]

T. Kovalenko$^1$, S. Vincent$^1$, V.I. Yukalov$^{1,2,}$\footnote{corresponding author}, 
and D. Sornette$^{1,3,4}$ } \\ [3mm]

{\it 
$^1$ETH Z\"urich, Department of Management, Technology and Economics, \\
Scheuchzerstrasse 7, 8092 Z\"urich, Switzerland \\ [2mm]

$^2$Bogolubov Laboratory of Theoretical Physics, \\
Joint Institute for Nuclear Research, Dubna 141980, Russia \\ [2mm]

$^3$Swiss Finance Institute, c/o University of Geneva, \\
40 blvd. Du Pont d'Arve, CH 1211 Geneva 4, Switzerland \\ [2mm]

$^4$Institute of Risk Analysis, Prediction and Management (Risks-X), \\
Southern University of Science and Technology, Shenzhen, China 
}
\end{center}

\vskip 1cm

\begin{abstract}
We present the first calibration of quantum decision theory (QDT) to a dataset of 
binary risky choice. We quantitatively account for the fraction of choice reversals 
between two repetitions of the experiment, using a probabilistic choice formulation 
in the simplest form without model assumption or adjustable parameters. The prediction 
of choice reversal is then refined by introducing heterogeneity between decision makers 
through their differentiation into two groups: ``majoritarian'' and ``contrarian'' 
(in proportion 3:1). This supports the first fundamental tenet of QDT, which models 
choice as an inherent probabilistic process, where the probability of a prospect can 
be expressed as the sum of its utility and attraction factors. We propose to parameterise 
the utility factor with a stochastic version of cumulative prospect theory (logit-CPT), 
and the attraction factor with a constant absolute risk aversion (CARA) function. 
For this dataset, and penalising the larger number of QDT parameters via the Wilks 
test of nested hypotheses, the QDT model is found to perform significantly better 
than logit-CPT at both the aggregate and individual levels, and for all considered 
fit criteria for the first experiment iteration and for predictions (second ``out-of-sample'' 
iteration). The distinctive QDT effect captured by the attraction factor is mostly 
appreciable (i.e., most relevant and strongest in amplitude) for prospects with big 
losses. Our quantitative analysis of the experimental results supports the existence 
of an intrinsic limit of predictability, which is associated with the inherent 
probabilistic nature of choice. The results of the paper can find applications both 
in the prediction of choice of human decision makers as well as for organizing the 
operation of artificial intelligence. 

\end{abstract}

\vskip 1cm

{\parindent=0pt
{\bf Keywords}: Quantum decision theory, Prospect probability, Utility factor, 
Attraction factor, Stochastic cumulative prospect theory, Predictability limit }

\newpage

\section{Introduction}

The life of every human being (and even of almost every alive being) is a permanent 
chain of decisions and actions resulting from these decisions. There are two types of 
decisions: Individual decisions taken by separate individuals without consulting others,
and collective decisions accepted after discussions with other involved individuals.
Humans are social animals and many their decisions are collective, being influenced
by social relations \cite{Perc_1,Perc_2,Jesup_3}. Nevertheless, the first step in 
developing any decision theory is the characterization of individual decision making.

There exist several variants of decision theory, whose peculiarities and limitations 
are discussed below. An original approach in decision theory, called quantum decision 
theory (QDT) has been advanced by the authors \cite{YSQDT08}. The idea of this approach 
is the use of techniques of quantum theory for describing the complex structure of 
realistic decisions containing the rational reasoning as well as irrational emotional
parts. It turns out that this rational-irrational duality can be successfully described
by quantum techniques developed for characterizing the theory of quantum measurements.
At the same time, mathematics of quantum theory is just a convenient tool not requiring 
that decision makers be in any sense quantum devices. After understanding the pivotal 
technical points of the approach, it is possible to reformulate the basic concepts so 
that it would be straightforward to employ the approach without resorting to quantum 
terminology. This concerns the main content of the present paper, whose reading does 
not need any knowledge of quantum theory. Quantum analogies and foundations are mentioned 
in Appendix and can be neglected by those who are not acquainted with quantum notions. 

The QDT approach can be applied to individual as well as to collective decision making,
when agents form a society and repeatedly exchange information between each other
\cite{Yukalov_4,Yukalov_5,Yukalov_6}. However, the ultimate aim is not merely qualitatively 
describe the novel approach in decision theory but to develop it to the level allowing
for its use in practical problems needing rather accurate quantitative predictions. This
paper is the first such an attempt of calibrating QDT, opening the ways for the following 
practical usage of the approach.        
    
The principal goal of decision theory is to understand and predict the choices 
of decision makers, in particular when the decisions involve risky options.  
``Classical'' economists use the {\it Homo economicus} assumption that decision 
making is the deterministic process of maximising an expected utility 
\cite{Bernoulli,vonNeumannMor47,Savage}. This formulation has been shown to lead 
to many paradoxes when confronted with real human decision makers.

Observed issues of ``classical'' models can be generalized into two classes:
\begin{itemize}
\item 
systematic deviations of behavior from predictions based on the expected utility, 
which led to a proliferation of behavioral models; 

\item 
choice variability over time, which gave rise to probabilistic extensions of 
deterministic models.
\end{itemize}

Systematic studies of behavioural patterns, revealed by accumulated empirical 
data, indicate violation of the classical axioms. These violations include: 
(a) common consequence and common ratio effects, which are inconsistent with 
the axiom of independence from irrelevant alternatives \cite{Allais53}; (b) the 
preference reversal phenomenon \cite{LichtensteinSlovic71,Lindman71} that is 
associated with a failure of procedure invariance and the axiom of transitivity 
\cite{LoomesSugden83}; and (c) framing effects as a breakdown of descriptive 
invariance \cite{TverskyKahneman81}. Many models have been introduced to explain 
and predict observed cognitive and emotional biases \cite{Camereretl03,Machina08}. 
A number of theories have been advanced, such as prospect theory 
\cite{Edwards55,Edwards62,KaTversky}, rank-dependent utility theory 
\cite{Quiggin82,Quiggin93}, cumulative prospect theory \cite{KaTversky2}, 
configural weight models \cite{Birnbaum74,Birnbaum-rank08}, regret theory 
\cite{LoomesSugden82,LoomesSugden87}, maximin expected utility model 
\cite{GilboaSchmeidler89}, Choquet expected utility model 
\cite{Gilboa87,Schmeidler89} and many others. However, various attempts to extend 
utility theory by constructing non-expected utility functionals do not avoid common 
pitfalls in modeling risk aversion \cite{SafraSegal08}, cannot in general resolve 
the known classical paradoxes such as the conjunction fallacy, disjunction effect, 
and were criticized for employing ambiguity aversion to rationalize Ellsberg 
choices \cite{Al-NajjarWein}. Moreover, extending the classical utility theory has 
been claimed ``ending up creating more paradoxes and inconsistencies than it 
resolves'' \cite{Al-NajjarWein}.

The observed variability of choice over time for one decision maker motivated 
the development of probabilistic extensions of deterministic ``classical" models. 
The need to prioritise the advancement of research concerned with probabilistic 
descriptions, as compared to the development of new versions of deterministic 
behavioural models, has been pointed out for example in 
\cite{HeyOrme94,Hey05,Rieskampreview08}. In fact, the axiomatic expected 
utility theory, when extended to incorporate truncated random errors, has 
been demonstrated to explain experimental data at least as well as cumulative 
prospect theory \cite{Blavatskyy05}. At the same time, the assumptions behind 
the stochasticity of choices have a wide range of interpretations, from erroneous 
and noisy execution to a useful evolutionarily feature, or left implicit. Moreover, 
different probabilistic specifications for the same core (deterministic) model 
have been shown to produce opposite predictions 
\cite{HeyOrme94,Hey95,LoomesSug95,CarboneHey00,Loomes05}. We review this topic 
in the next section.

Thus, modifications of ``classical'' models, by incremental additions of 
behavioral parameters and stochastic elements, had led to an impressive growth 
of the literature and of its complexity, without however convergence towards a 
commonly accepted solution for the two classes of paradoxes.

In the last decade, there has been a growing interest in a conceptually new way 
of modeling decisions by employing a toolbox that was originally developed for 
quantum mechanics. Within the ``quantum'' approach, decision making is seen as 
a process of deliberation between interfering choice options (prospects) with 
a probabilistic result, i.e. a probabilistic decision. Thus, it provides a 
parsimonious explanation for both modeling issues: systematic deviations from 
a rational choice criterion considered in isolation appear, unconsciously or 
intentionally, due to the presence and certain formulation of interconnected 
prospects. And the observed choice stochasticity is a manifestation of the 
inherently probabilistic nature of decision making.

The factors causing interference effects in decision processes include subjective 
and subconscious processes in the decision maker's mind associated with available 
prospects coexisting with the prospect(s) under scrutiny for a decision action. 
This includes memories of past experiences, beliefs and momentary influences. All 
these operations in the mind of the decision maker may contribute to the existence 
of interferences between the different prospects and/or between a given prospect 
and his/her state of mind. In the Appendix summarising quantum decision theory, 
such interference effects are quantified by the {\it attraction factor}, which is 
one of the main objects of quantitative investigation in the present work.

The quantum decision theory that we follow here was first introduced in Ref.
\cite{YSQDT08}, with the goal of establishing a holistic theoretical framework 
of decision making. Based on the mathematics of Hilbert spaces, it provides a 
convenient formalism to deal with (real world) uncertainty and employs non-additive 
probabilities for the resolution of complex choice situations with interference 
effects. The use of Hilbert spaces constitutes the simplest generalization of the 
probability theory axiomatized by \cite{Kolmogorov56} for real-valued probabilities 
to probabilities derived from algebraic complex number theory. By its mathematical 
structure, quantum decision theory aims at encompassing the superposition processes 
occurring down to the neuronal level. This becomes especially important for 
composite (uncertain) measurements, with a formulation that differs from the 
diverse forms of probabilistic choice theory, including random preference models 
(mixture models), as the summary presentation of quantum decision theory in the 
Appendix should help comprehend. Numerous behavioural patterns, including those 
causing paradoxes within other theoretical approaches, are coherently explained 
by quantum decision theory 
\cite{YSQDT08,YSentropy,YSmathQDT,YSQDT1ThDec11,YSQDT14cogn,YSPosOp,
YSQDT15Philtrans,YSQDT15frontpsy}.

There are several alternative versions of quantum approach to decision making, 
which have been proposed in the literature, as seen for instance with the 
books \cite{Khren10,BuseBruz12,Bagarello13} and the review articles 
\cite{YSentropy,Sorrev2014,BusewangKh14,Ashtianiza15}, where citations to the 
previous literature can be found. The version of Quantum Decision Theory 
(henceforth referred to as QDT), developed in 
Refs. \cite{YSQDT08,YSentropy,YSmathQDT,YSQDT1ThDec11,YSQDT14cogn,YSPosOp,
YSQDT15Philtrans,YSQDT15frontpsy} and used here, principally differs from all 
other ``quantum'' approaches in two important aspects. First, QDT is based on a 
self-consistent mathematical foundation that is common to both quantum measurement 
theory and quantum decision theory. Starting from the theory of 
quantum measurements of von Neumann \cite{vonNeumann55}, the authors have generalized it to the case of uncertain or 
inconclusive events, making it possible to characterize uncertain measurements and 
uncertain prospects. Second, the main formulas of QDT are derived from general 
principles, giving the possibility of general quantitative predictions. In a series 
of papers \cite{YSQDT08,YSentropy,YSmathQDT,YSQDT1ThDec11,YSQDT14cogn,YSPosOp,
YSQDT15Philtrans,YSQDT15frontpsy} the authors have compared a number of predictions 
with empirical data, without fitting parameters 
\cite{YSQDT1ThDec11,YSQDT14cogn,YSQDT15Philtrans,YSQDT15frontpsy}. This is in contrast 
with the other ``quantum approaches'' by other researchers consisting in constructing 
particular models for describing some specific experiments, with fitting the model 
parameters from experimental data.

Until now, predictions of QDT were made at the aggregate level, non 
parametrically and assuming no prior information. The present study intends 
to overcome these limitations, by developing a first parametric analytical 
formulation of QDT factors, enlarging the area of practical application of 
the theory and enabling higher granularity of predictions at both aggregate 
and individual levels. 

For the first time, we engage QDT in a competition with decision making 
models, based on a mid size raw experimental data set of individual choices. 
The experiment was iterated twice (henceforth referred to as time 1 and time 2) 
and consists of simple choice tasks between two gambles with known outcomes and 
corresponding probabilities (i.e. binary lotteries). The data analysis reveals 
an inherent choice stochasticity, adding to the existing evidences, and supporting 
the probabilistic approach of QDT. 

As a classical benchmark, we consider a stochastic version of cumulative prospect 
theory (henceforth referred to as logit-CPT) that combines cumulative prospect 
theory (CPT) with the logit choice function. Note that other models associated 
with ``classical'' theories, such as expected value and expected utility theory, 
are nested within it. For review on tests of nested and especially non-nested 
hypotheses, see \cite{GourierouxMonfort94}. 

Within QDT, a decision maker, who is exposed to several options, can choose any 
of these prospects with a certain probability. Thus, each choice option is 
associated with a prospect probability, which can be calculated as a sum of two 
factors: utility and attraction. In this paper, for the parametric formulation 
of QDT, we adopt the stochastic CPT approach (logit-CPT) for the utility factor, 
and incorporate a constant absolute risk aversion (CARA) into the attraction 
factor. This allows us to separate aversion to extreme losses and transfer it 
into the attraction factor.

We estimate parameters of the logit-CPT model and the utility factor of our QDT 
model with the hierarchical Bayesian method, as implemented in 
\cite{NilssonRiesWagen11,ScheibehennePa15,HMLMurphy} and in \cite{Farrell,poissonPDF}, 
using identical data set as \cite{HMLMurphy}, which ensures straightforward model 
selection. The proposed QDT formulation is found to perform better at both aggregate 
and individual levels, and for all considered criteria of fit (time 1) and prediction 
(time 2). As expected, the most noticeable effect is achieved for prospects involving 
large losses, whereas the overall improvement is small on average.

The difficulty of achieving significant improvements in the prediction of human 
decisions, despite persistent attempts of different approaches, raises the question 
of the limit of predictability. We propose to rationalize quantitatively the limits 
of predictability of human choices in terms of the inherent stochastic nature of 
choice, which implies that the fraction of correctly predicted decisions is also 
a random variable. We thus propose a theoretical distribution of the individual 
predicted fractions, and compare it successfully to the experimental results.

The main contributions of this paper are the following. Analysing a previously 
studied experimental data set comprising 91 choices between two lotteries presented 
in random order made by 142 subjects repeated at two separated times, we suggest 
an original quantification of the choice reversals between the two repetitions.
This provides a direct support for one of the hypotheses at the basis of QDT that 
decision making may be intrinsically probabilistic. Our formulation gives a very 
intuitive grasp of how the probabilistic component of decision making can be revealed.
Our second contribution is to propose a simple efficient parameterisation of QDT that 
is used to calibrate quantitatively the experimental data set. This extends previous 
tests of QDT made at the population level, for instance focusing on the verification 
of the quarter law of interference. The proposed parametric analytical formulation 
of QDT combines elements of a stochastic version of Cumulative prospect theory (logit-CPT) 
for the utility factor $f$, and  constant absolute risk aversion (CARA) for the 
attraction factor $q$. One important insight is that the level of loss aversion inverted 
from QDT is significantly smaller than the loss aversion inferred from the benchmark 
logit-CPT implementation, suggesting that interference effects accounted by the QDT 
attraction factor provide a better explanation of empirical choices. The horse-race 
between the QDT model and the reference classical logit-CPT model is clearly won by 
the former at both aggregate and individual levels, and for all considered criteria.
Finally, QDT uncovers an accentuation of the aversion to extreme losses as embodied 
by the QDT attraction factor, which is responsible for noticeable improvement of the 
calibration of the model for mixed and pure loss lotteries involving big losses. 

The article thus aims to bridge traditional and quantum(-like) decision theories, 
and to contribute to their comparison along the two introduced threads:
(i) systematic deviations from classical axioms, i.e. significance of a 
quantum-interference effect (in Section \ref{sec:QDT}), which is embedded in 
a broader discussion on (ii) the interpretation of choice stochasticity, i.e. 
the implications of a pure probabilistic nature of choice (other Sections).
This is done by the following structure of the paper. Section \ref{Review_stochas} 
is an overview of stochastic decision models and alternative interpretations of the 
nature of choice variability. Section \ref{uymjt} presents empirical evidence 
supporting probabilistic choice frameworks. A simple nonparametric probabilistic 
model is proposed that can predict the frequency of preference reversals on the 
basis of the observed fraction of individuals making a choice in the first 
iteration of the experiment. Section \ref{sec:QDT} compares calibration and 
prediction results of the QDT model with the ones obtained for the stochastic 
model of CPT, both at the aggregate and individual levels. Section \ref{yjnheew} 
investigates the limits of the improvement of choice predictions in the presence 
of the proposed probabilistic nature of decision making. Section \ref{etmyjmk,u5rj} 
develops a link between the probabilistic shift model and QDT, and 
Section \ref{sec:ccl} concludes.

\section{Stochastic decision models and the nature of choice variability: 
from ``error'' to ``evolutionary advantage''} \label{Review_stochas}

One of the difficulties in modeling decision makers' behaviour is associated with 
the variability of their choices. There is compelling evidence from a substantial 
body of psychological and economic research that people are not only different in 
their preferences (corresponding to between-subject variability), but, importantly, 
they do not perform deterministic choices (and thus exhibit within-subject 
variability) \cite{MostellerNogee51,Tversky69,Hey01}. A person in a nearly identical 
choice situation on repeated occasions often opts for different choice alternatives, 
and the magnitude of choice probability variations is context dependent. Choice 
reversal (switching) rate has been reported between 20 and 30\%, and for some tasks 
can be close to 50\% 
\cite{Camerer89,StarmerSugden89,HeyOrme94,BallingerWil97,RieskampBuseMel06,
Regenwetteretal11}. Thus, at the aggregate and individual levels, decision makers 
do not seem to settle on the choice that exhibits the largest unequivocally defined 
desirability. To account for variability of individual choice, and to help formalise 
economic models, the previously mentioned (expected utility and non-expected utility) 
deterministic theories have been combined with stochastic components. 

At an early stage, the development of probabilistic models of choice and preference 
was associated with psychophysics. Thurstone's law of comparative judgement 
\cite{Thurstone27} and Luce's choice axioms \cite{Luce59} imply models that are 
specimens of the two broad classes of probabilistic choice models. For historical 
connections between Thurstonian model and Luce's choice model, see for example 
\cite{Pleskac12}. Respectively, the classes are 
\cite{LuceSuppes65,Marley92,RieskampBuseMel06}:
(i) random utility models, which combine stochastic utility function with 
deterministic choice rule, i.e. the maximisation of a random utility at each 
repetition of a decision; and (ii) constant (fixed) utility models, which assume 
a fixed numerical utility function over the choice outcomes complemented by a 
probabilistic choice rule, i.e. response probabilities that are dependent on the 
scale values of the corresponding outcomes. For instance, cumulative prospect 
theory has been supplemented with the probit \cite{HeyOrme94} or the logit 
choice functions \cite{CarboneHey95,BirnbaumChavez97}. Another class of models 
suggests the existence of (iii) a random strategy selection (or random preferences) 
such that, within each strategy (or preference state), both elements, utility and 
choice process, are deterministic. Random preference models (aka mixture models) 
assume probabilistic distribution of decision maker's underlying (latent) 
preferences, and interpret choices as if they are observations drawn from such 
a distribution 
\cite{NiedereeHey89,NiedereeHey92,NiedereeHey97,Regenwetter96,RegenwetterMar01,
Regenwetteretal10,Regenwetteretal11,LoomesPogrebna14,LoomesPogrebna17}.
Different stochastic specifications have been explored, and a large literature 
has evolved 
\cite{Marschak60,BlockMars60,Yellott77,IversonFal85,HeyerMau87,Marley68,Marley89a,
Marley89b,LuceNarens94,HarlessCam94,Hey95,HeyCarbone95,Luce95,BallingerWil97,
LoomesSug95,LoomesSug98,McFaddenTrain00,Fishburn01,Loomesetal02,Hey05,Myungetal05,
Birnbaum06,Rieskampreview08,Wilcox08,Davis-Stober09,BlavatskyyPo10,Conteetal11,
Regenwetteretal12,Regenwetteretal14,MasNax16}.

Summarising the above, the necessity of a stochastic approach for the modeling 
of choices is widely recognized. At the same time, we suggest that assumptions 
about the {\it nature of the stochasticity of choices} deserve particular 
attention, and some of the current interpretations may require reconsideration. 

Firstly, one of the prevalent views in the literature is that the observed 
probabilistic choices are a result of the bounded rationality of decision makers. 
Empirically documented effects, such as preference reversal, similarity, compromise 
and attention effects, have often been classified as ``inconsistencies'' of people's 
behaviour \cite{RieskampBuseMel06}, which is mistaken and noisy \cite{Hey05}. In 
this interpretation, the core of the choice process is still deterministic, in the 
sense that the decision maker strives to choose the best alternative but, doing so, 
he/she makes errors either in the evaluation of the options(e.g. a measurement 
error \cite{HeyOrme94}) or in the implementation of his/her choice (e.g. an application 
error with a constant probability of its occurrence \cite{HarlessCam94,RieskampOtto06}). 
The standard way of using such a stochastic approach is to assume a probability 
distribution over the values characterizing the errors made by the subjects in the 
process of decision making. Such stochastic decision theories can be termed as 
``deterministic theories embedded into an environment with stochastic noise", and 
are typical of (i) random utility models and (ii) fixed utility models.

Another perspective is to consider that the stochastic elements are technical 
devices added to the deterministic theory to allow for its calibration to 
experiments, with the implicit or explicit understanding that the stochastic 
component of the choice may result from the component of the utility of a decision 
maker that is unknown or hidden to an observer trying to rationalize the choices 
made by the decision maker \cite{LuceSuppes65,McFadden74}. This interpretation is 
relevant to models with (iii) random preferences. In this view, a probabilistic 
model accounts for the empirically observed behavioural inconsistencies, however 
their origin and causes are often put out of the scope of the discussion.

Finally, stochastic assumptions often remain implicit, though they play a 
defining role in the formulation of testable hypotheses and the selection 
of methods of statistical inference \cite{Hey05}. Different probabilistic 
specifications have been shown to lead to possibly opposite predictions for 
the same core (deterministic) theory 
\cite{HeyOrme94,Hey95,LoomesSug95,CarboneHey00,Loomes05}. These emphasize that 
``stochastic specification should not be considered as an `optional add-on,' but 
rather as integral part of every theory which seeks to make predictions about 
decision making under risk and uncertainty'' (p. 648) \cite{LoomesSug95}.

In our view, strong probabilistic theories, which assign a precise probability 
for each option to be chosen, provide valuable modeling tools. They should not 
be perceived as mere extensions of deterministic core theories. Rather, a general 
probabilistic framework that highlights the intrinsic stochastic origin of decision 
making should be put to the forefront.

Arguably, among the classes named above, random preference models (mixture models) 
correspond the most to this approach \cite{Loomes15}.

For example, models based on stochastic processes have been introduced to 
represent mental deliberation and account for choice and reaction time jointly, 
as well as to model (longitudinal) panel data. These include decision field theory 
\cite{BuseTown93}, ballistic accumulator models \cite{BrownHeat08}, media theory 
\cite{Falmagne96,FalmagneOv02}, sequential sampling models \cite{Forstmannrawag16}, 
stochastic token models of persuasion \cite{Falmagne97} and so on. 

The quantum decision approach that we will present and test here resonates with 
this strand of research emphasizing that decision making might be intrinsically 
probabilistic. While there is a huge literature briefly mentioned above on 
probabilistic decisions, the prominent advantage of quantum decision theory 
is that it is by essence structurally probabilistic. In other words, the whole 
theoretical construction of how people make decisions cannot be separated from 
a probabilistic frame. Contrary to classical stochastic decision theory in 
economics, we do not assume that choices are deterministic, with just some weak 
disturbance associated with errors. In quantum decision theory, a probabilistic 
decision is not a stochastic decoration of a deterministic process: a random 
part is unavoidably associated with any choice, which can be interpreted as 
representing subconscious hidden neuronal processes. 

The difference between the classical stochastic decision theory in economics 
and quantum decision theory is similar to the difference between classical 
statistical physics and quantum mechanical theory. In the former, all processes 
are assumed to be deterministic, with statistics coming into play because of 
errors and statistical fluctuations, such as no precise knowledge of initial 
conditions and the impossibility of measuring exactly the locations and velocities 
of all particles. In contrast, quantum mechanics postulates that the precise 
states of particles are unknowable and, in the standard so-called Copenhagen 
interpretation, inherently so due to the essence of the laws of Nature. Similarly, 
the quantum decision theory used here embraces the view and actually requires in 
its very construction that decision making is intrinsically probabilistic. 

There is a growing perception that the existence of probabilistic choices can be 
actually optimal in a certain broader sense. For instance, the occasional selection 
of alternatives that are dominated according to a particular desirability criterion, 
can actually be beneficial for an individual and/or a group when measured over large 
time scales. In evolutionary biology, a long-term measure of utility is known as 
reproductive value, which represents the expected future reproductive success of 
an individual. Natural selection favors those individuals, who behave as if 
maximising their reproductive value \cite{HoustonMcNamara99}. Similarly, traits 
such as ``strong cooperation'' \cite{Henrich04} and ``altruistic punishment'' 
\cite{FehrGachter00,FehrGachter02,FehrFischbacher03} are costly to the individual 
and do not seem to make sense from the perspective of a person's utility 
maximisation, but are selected in evolutionary agent-based models of competing 
groups in stochastic environments \cite{HetzerSor13a,HetzerSor13b}.

Stochastic decision making can provide an evolutionary advantage by being 
instrumental in overcoming adverse external and internal factors by:
\begin{itemize}
\item 
exploring uncertain complex environments with unknown feedbacks;

\item 
discovering available choice options and variations of their utilities over time 
\cite{McNamaraetal14};

\item 
refining preferences by sampling and through comparative judgment 
\cite{Stewartetal06};

\item learning using ``trials and errors'' and bridging a ``description-experience 
gap'' \cite{HertwigErev09};

\item adapting strategies at an individual and group levels, and introducing 
diversification.
\end{itemize}

Thus, choice variability should not be considered as an anomaly or exception. On 
the contrary, it may be an advantageous trait developed in humans, whose evolution 
is linked to a stochastic and uncertain environment. This view, incorporating the 
evidences reported in this paper, has been recently briefly summarised 
in \cite{Sornette17}.

\section{Empirical evidence supporting probabilistic choice formulations}
\label{uymjt}

\subsection{Basic experimental setting}
\label{subsec:expdesign}

Choice between gambles was called ``the fruit fly of decision theory'' 
\cite{KahnemanTversky00} as one of the simplest settings of choice under risk 
and elicitation of risk preferences. We consider a choice between two gambles $A$ 
and $B$ (i.e. binary lotteries), each of which consists of two outcomes, in a range 
from $-100$ to $100$ monetary units (MU), with known probabilities that sum to one, 
as shown in Table \ref{tab:gamble}. Participants had to choose one of the lotteries, 
and were not allowed to express either indifference or lack of preference, thus a 
two-alternative forced choice (2AFC) paradigm was implemented. The experimental 
set included 91 pairs of static lotteries (i.e. outcomes and probabilities were 
not contingent upon a preceding choice of a decision maker) of four types: 35 pairs 
of lotteries with gains only; 25 pairs with losses only; 25 pairs of mixed lotteries 
with both gains and losses; and 6 pairs of mixed-zero lotteries with one gain and 
one loss and zero (status quo) as the alternative outcome. The first three types 
of binary lotteries cover the spectrum of risky decisions, while the mixed-zero 
type allows for measuring loss aversion separately from risk aversion 
\cite{Rabin00,Wakker05}. The set of lotteries was compiled from lotteries 
previously used in \cite{HoldLaury02,Gaechteretal07,Rieskampreview08}. The 
collected empirical data of 142 participants (from the subject pool at the Max 
Planck Institute for Human Development in Berlin) was obtained from \cite{data}.
 
Additional details of the experimental design, including a complete list of binary 
lotteries, can be found in \cite{HMLMurphy}, which exploits the same data set in 
their calibration of stochastic cumulative prospect theory (logit-CPT).

\begin{table}[h]
\caption{\small Choice between two finite valued lotteries. If a decision maker chooses 
lottery $A$, then the outcome will be $V^A_1$ with probability $p^A_1$, and $V^A_2$ 
with probability $p^A_2=1-p^A_1$, and similarly if he/she chooses lottery $B$ with the 
superscript changed from $A$ to $B$. The outcomes can be either positive (gains) 
or negative (losses).}
\label{tab:gamble}
\renewcommand{\arraystretch}{1.25}
\centering\begin{tabular}{l |c r}
 &Outcomes $\;\&$ Probabilities &\\
Lottery $A$ & $(V^A_1 ; p^A_1)$  or $(V^A_2 ; p^A_2)$ & $p^A_2=1-p^A_1$\\
\hline
Lottery $B$ & $(V^B_1 ; p^B_1)$  or $(V^B_2 ; p^B_2)$ & $p^B_2=1-p^B_1$
\end{tabular}
\end{table}

The experiment was repeated twice at an approximately two weeks interval 
(henceforth referred to as time 1 and time 2) with the same 142 subjects and the 
same set of 91 binary lotteries. At time 1, the order of lottery items and their 
spatial representation within a pair was randomized, and displayed in the reverse 
order at time 2. By ``spatial representation within a pair'', we refer to a 
presentation as in Table 1 where one lottery is presented as lottery A and the second 
of the pair is called lottery B. But the same pair could be arranged in the opposite 
order where the first presented lottery is B and the second one is A. Consequently, 
the order and presentation effects were mitigated. The experiment was incentive 
compatible with a two-part remuneration: a fixed participation fee, and a varying 
payment based on a randomly selected lottery from the choice set, which was played 
out at the end of both experimental sessions.

The recording of the choices between the same alternatives by the same subjects 
at two different times allows one to perform in-sample modeling (at time 1) and 
out-of-sample predictions (of time 2).

\subsection{Analysis of the consistency and differences between times 1 and 2}

\subsubsection{Stability of the aggregate choice frequencies and variability of 
the individual preferences}

\begin{figure*}[ht]
\centering
\includegraphics[width=0.6\textwidth]{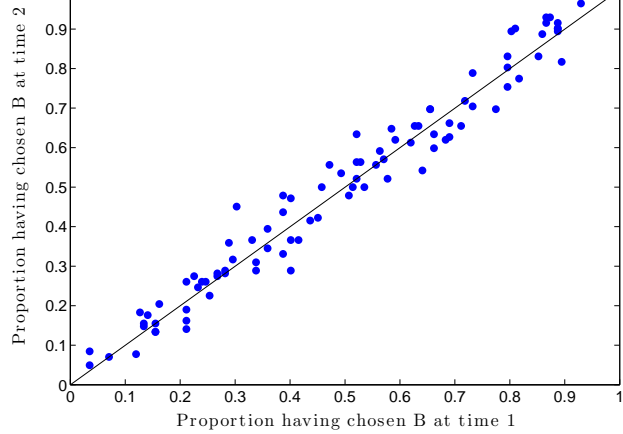}
\caption{\small
Proportion of decision makers having chosen option $B$ at time 2 as a 
function of the proportion of decision makers having chosen option $B$ 
at time 1 (there are 91 points, one for each of the 91 presented pairs 
of lotteries).
}
\label{fig1}
\end{figure*}

Figure \ref{fig1} compares the proportion of decision makers among the 142 
subjects who chose option $B$ at both time 1 and time 2 for each of the 91 binary 
lotteries. We refer to this proportion as the experimental ``frequency'' of choice 
$B$ in a given pair of lotteries. As the diagonal in Figure \ref{fig1} 
represents what would be a perfect reproducibility of the choices at the two times, 
at the aggregate level, the first overall observation is that the frequency of the 
choice in each pair of lotteries is rather stable from time 1 to time 2, since the 
data points tend to cluster along the diagonal. The linear relationship shows that 
decision makers, as a group, exhibit a stable preference across time. The fact that 
the lotteries sample essentially the full frequency interval $[0,1]$ confirms that 
they cover a large set of preferences, from obvious gambles where one of the 
prospects is almost always preferred to more ambivalent gambles. The frequencies 
of the choices shown in Figure \ref{fig1} is a manifestation of the type 
of choices.

Stability over time of aggregate preferences is confirmed by the analysis of the most 
common choice, i.e. a lottery within each pair that is chosen by the majority of 
subjects. For this dataset, only for 4 out of 91 lottery pairs the most common 
choice shifted between time 1 and time 2. These lottery pairs are listed in Table 
\ref{tab:maj_choice_shift}. Notably, choice alternatives within each of these 
4 pairs are characterized by relatively close expected values. 

\begin{table}[ht]
\renewcommand{\arraystretch}{1.25}
\centering
\caption{\small The only 4 out of 91 pairs of lotteries, for which the most common 
(i.e. majority) choice has shifted between two repetitions of the experiment 
(time 1 and time 2). The choice is between lottery A and B. The most common 
choice at time 1 is highlighted in bold. Choice alternatives within each pair 
are characterized by relatively close expected values.}
\label{tab:maj_choice_shift}
\begin{tabular}{@{}cccclcccclcc@{}}
\multicolumn{4}{c}{Lottery $A$}                            &  & \multicolumn{4}{c}{Lottery $B$}                            &  & \multicolumn{2}{c}{Expected value} \\ \hline
$V_1^A$     & $p_1^A$       & $V_2^A$      & $p_2^A$       &  & $V_1^B$     & $p_1^B$       & $V_2^B$      & $p_2^B$       &  & Lottery $A$      & Lottery $B$     \\  \hline
\textbf{56} & \textbf{0.05} & \textbf{72}  & \textbf{0.95} &  & 68          & 0.95          & 95           & 0.05          &  & 71.2             & 69.35           \\
\textbf{88} & \textbf{0.29} & \textbf{78}  & \textbf{0.71} &  & 53          & 0.29          & 91           & 0.71          &  & 80.9             & 79.98           \\
\textbf{-8} & \textbf{0.66} & \textbf{-95} & \textbf{0.34} &  & -42         & 0.93          & -30          & 0.07          &  & -37.58           & -41.16          \\
96          & 0.61          & -67          & 0.39          &  & \textbf{71} & \textbf{0.50} & \textbf{-26} & \textbf{0.50} &  & 32.43            & 22.5            \\ \hline
\end{tabular}
\end{table}

\begin{figure*}[ht]
\centering
\includegraphics[width=0.6\textwidth]{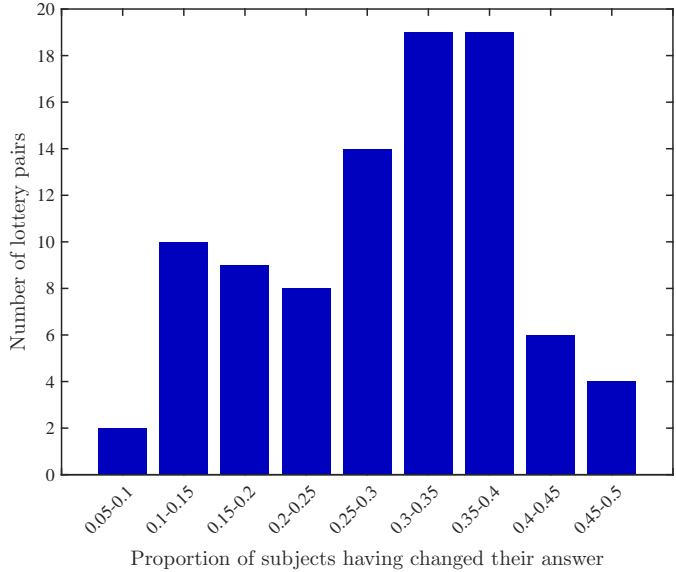}
\caption{
\small
Histogram over all 91 lottery pairs of the proportion of decision makers 
having changed their choice between times 1 and 2. Note that the ordinate 
values of the ten bins sum up to 91. For more than half of the considered 
lottery pairs (48 out of 91), more than 30\% of the subjects shifted their 
preference from $A$ to $B$ or vice-versa, between times 1 and 2.
}
\label{fig2}
\end{figure*}

Stability at the aggregate level is accompanied by variability of individual 
choices. In Figure \ref{fig1}, a significant scatter around the diagonal 
indicates a stochasticity in the revealed preferences of the decision makers. 
The individual deviation of choices between times 1 and 2 is further quantified 
in Figure \ref{fig2}, which plots the number of lottery pairs for which 
a given proportion of subjects has changed their choice. One can observe that 
individual choices of decision makers may vary significantly over time. In more 
than half of the binary lotteries, more than 30\% of the subjects changed their 
answer between time 1 and time 2. The average rate of choice reversal (switching) 
per subject is slightly higher than 29\%, which is in line with the values previously 
reported in the literature.

\subsubsection{Quantitative rationalisation via probabilistic choices
\label{shifthomog}}

The combined observation of the overall stability of the choices at the 
aggregate level (Figure \ref{fig1}) and their variability at the individual 
level (Figure \ref{fig2}) adds to the large body of empirical literature 
discussed in the introduction that purports that decisions are probabilistic rather 
than deterministic. However, it is interesting to test it quantitatively as follows. 
For this, we propose a non-standard approach, which abstracts from any assumption 
on the probability model, algebraic core, and on the stimuli that promote the 
decisions. We straightforwardly derive the probability of a choice shift between 
times 1 and 2 from a single ingredient -- the frequency of that choice observed at 
time 1 as a measure of the corresponding prospect probabilities, without any fit.
In other words, the frequency of a given choice over the population of decision 
makers is taken as a probe for the underlying probability for that choice, used in 
the usual frequentist interpretation of probabilities \cite{Kendall49}.

Considering a given pair of lotteries, let us denote by $X_t$ the event 
\emph{``choosing lottery $X\in\{A,B\}$ at time $t\in\{1,2\}$''}. For instance, 
if the decision maker chooses lottery $A$ at time 1 and lottery $B$ at time 2, 
this is represented by the combined event $A_1\bigcap B_2$. The overall stability 
of the choices at the aggregate level (figure  \ref{fig1}) suggests the 
parsimonious assignment of a fixed stable probability $p_j$ for each of the two 
choices in a given lottery pair $j$:
\begin{equation}
\mathbb{P}\left(A_{1,j}\right)=\mathbb{P}\left(A_{2,j}\right)=p_j
\label{htjye1}
\end{equation} 
and 
\begin{equation}
\mathbb{P}\left(B_{1,j}\right)=\mathbb{P}\left(B_{2,j}\right)=1-p_j~.
\label{htjye2}
\end{equation} 
This hypothesis consists in neglecting any heterogeneity between decision 
makers, thus assuming that they all have the same preference. Notwithstanding 
its simplicity, we now show that it is remarkably powerful at accounting for 
most of the observed shifts between times 1 and 2.

Indeed, because each choice among two lotteries within a pair is assumed 
probabilistic, this implies that repeating the experiment is expected to give 
possible choice shifts from $A$ to $B$ and vice-versa, just from the hypothesised 
probabilistic nature of the choice. Thus, the probability that a decision maker 
shifts her choice in a pair of lotteries is given by:
\begin{equation}
\mathbb{P}\left(\text{shift}\right)=
\mathbb{P}\left(A_1\bigcap B_2\right)+\mathbb{P}\left(B_1\bigcap A_2\right)~.
\label{rthnhba}
\end{equation}
This expression conveys the fact that the shift could occur from the choice 
$A$ at time 1 followed by the choice $B$ at time 2. This is represented by 
$A_1\bigcap B_2$. Or the decision maker might have chosen $B$ at time 1 followed 
by the choice $A$ at time 2. This is represented by $B_1\bigcap A_2$. Considering 
both scenarios together leads to expression (\ref{rthnhba}).

The analysed experiment was conducted twice with the same decision makers, facing 
the same set of lottery pairs. Therefore, the successive decisions $A_1\bigcap B_2$ 
or $B_1\bigcap A_2$ are dependent because it is a repeated measure by design. 
However, let us assume that, when they form their choice at time 2, decision 
makers have forgotten their choices performed at time 1 (which is likely in the 
experimental set-up as the two iterations -- time 1 and time 2 -- were conducted 
approximately 2 weeks apart and the choice orders have been randomised). In the 
framework where their decisions are solely and completely captured by equations 
(\ref{htjye1}) and (\ref{htjye2}) expressing an intrinsic probabilistic choice 
structure, for a pair of lotteries we have $\mathbb{P}\left(A_1\bigcap B_2\right)=
\mathbb{P}\left(B_1\bigcap A_2\right)=p(1-p)$, yielding
\begin{equation}
\mathbb{P}\left(\text{shift}\right)=2p\left(1-p\right)~.
\label{eq:P_shift_homo}
\end{equation}

This expression is the simplest and most parsimonious prediction for the probability 
$\mathbb{P}\left(\text{shift}\right)$ that a decision maker shifts her choice from 
time 1 to time 2. It is based on considering human behavior at the aggregate level, 
i.e., specifically that the fraction of persons making a given decision is equal 
(and equivalent) to the probability of a single random person to make that decision. 
This hypothesis is at the foundation of quantum decision theory and we refer to the 
Appendix and references therein for its motivation and justification. The second 
assumption underlying expression (\ref{eq:P_shift_homo}) is that decision makers have 
not kept the memory of their previous decision performed two weeks earlier. Given 
the neutral nature of the decisions (choosing between lotteries), this is likely to 
be a reasonable assumption. In the end, these assumptions made for simplicity have 
the virtue of leading to a prediction for $\mathbb{P}\left(\text{shift}\right)$ that 
is a function of a single variable $p$, which is itself measurable in the first 
experiment at time 1, leading to a parameter-free prediction.

In order to test the validity of prediction (\ref{eq:P_shift_homo}) on the 
experimental data, as mentioned above, we assume that the frequency of the most 
common choice for a given lottery pair over the ensemble of all decision makers 
(i.e. majority choice) is a proxy for the probability $p_j$.

Indeed, the frequency of the most common choice for a given pair $j$ of lotteries 
gives an estimate of the so-called frequentist definition of the corresponding 
probability \cite{Kendall49}, which converges to the true probability, if it 
exists, in the limit of very large samples. Similarly, we identify the probability 
$\mathbb{P}\left(\text{shift}\right)$ of a choice shift between times 1 and 2 with 
the proportion of decision makers having changed their choice between times 1 and 2. 

\begin{figure*}[ht]
\centering
\includegraphics[width=0.6\textwidth]{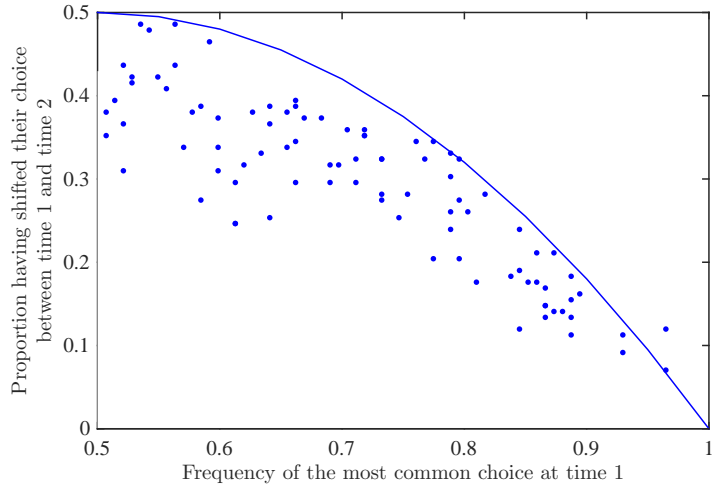}
\caption{\small
Proportion of decision makers having shifted their choice between time 1 
and time 2 as a function of the proportion choosing the most frequently 
chosen option at time 1 (there are 91 points, each one represents a pair 
of lotteries). The solid line represents the proportion of shifts predicted 
by a simple model (\ref{eq:P_shift_homo}), assuming homogeneity of preferences, 
their stability at the aggregate level and choice independence between times 
1 and 2. We stress that the solid line is not a ``fit'' as there are no 
adjustable parameters.
}
\label{fig3}
\end{figure*}

This prediction (\ref{eq:P_shift_homo}), which has no adjustable parameters, is 
shown as the blue smoothed continuous curve in Figure \ref{fig3}, 
which plots the proportion of decision makers having changed their choice between 
times 1 and 2 as a function of the frequency of the most common choice at time 1.

Figure \ref{fig3} shows that the main dependence is rather well 
captured by prediction (\ref{eq:P_shift_homo}), which we stress again is not a 
``fit'' as there is no adjustable parameter. Expression (\ref{eq:P_shift_homo}) 
has a simple intuitive interpretation: clear-cut choices associated with large 
$p_j$'s are aligned with strong and well-defined preferences, so that it is 
quite unlikely that a decision maker will change her choice; in contrast, when 
the frequency at time 1 for choosing a given lottery is close to even between 
the two lotteries, the decision makers are very likely to shift their choice at 
time 2. While these tendencies are obvious, what is less evident is the fact that 
the simple logical step leading to expression (\ref{eq:P_shift_homo}) accounts 
surprisingly well for the data, with no adjustment.

\subsubsection{Evidence of heterogeneity between decision makers: a parsimonious 
description \label{neytbgww}}

While the agreement between data and prediction shown in Figure \ref{fig3} is 
remarkable, given that the prediction has no adjustable parameters, it is also 
clear that the model over-estimates the number of decision shifts as the data tends 
to be systematically below the theoretical prediction, in particular for the pairs 
of lotteries with close ties, i.e. for which decision makers show a large heterogeneity 
of choices and the proportion choosing the most frequently chosen lottery is not much 
above 50\%. More precisely, for more frequently chosen options (with frequency of the 
most common choice above 75\%), the observed frequencies are closer to the theoretical 
prediction, while, for less frequently chosen options, the deviation is larger. This 
can explain the bimodal structure of the histogram in Figure \ref{fig2}. 

In order to arrive at prediction (\ref{eq:P_shift_homo}), we have used two main 
assumptions: 

(i) the choices between times 1 and 2 are made as if a single probability 
describes each of them (i.e. stability of the preferences and independence of 
choice between two repetitions of the experiment) and (ii) the decision makers' 
preferences are homogenous, so that the same single probability $\{p_j,\;j=1,...,91\}$ 
for each of the 91 pairs of lotteries characterises the full set of 142 subjects. We 
propose to keep the first assumption as part of a minimalist approach. As discussed 
briefly above, the second assumption flies in the face of enormous empirical evidence 
supporting the proposition that human decision makers exhibit significantly different 
risk preferences. This is particularly relevant to our discussion since the choices 
between the pairs of lotteries are specifically sensitive to the different levels of 
risk (as well as payoffs) associated with the competing lotteries in each pair. 

Relaxing the assumption that all decision makers are identical can immediately be 
seen to help removing the discrepancy observed in Figure \ref{fig3}.
Indeed, consider the simplest situation generalising heterogeneity, which consists 
in assuming the presence of two groups  $i\in\left\{1,2\right\}$ of decision makers 
of size $142 F$ and $142 (1-F)$, respectively (with $0 < F < 1$), for which 
$P(A^1_{1,j}) = P(A^1_{2,j}) = p_{1,j}$ and $P(A^2_{1,j}) = P(A^2_{2,j}) = p_{2,j}$,
where $A^i_{t,j}$ is the most frequent choice at time $t$ in a lottery pair $j$ by 
group $i$, and $p_{i,j}$ is a corresponding fixed probability of that choice for 
group $i$. Thus, for a given lottery pair the aggregate (over both groups) choice 
probability is
\begin{equation}
p=p_1F+p_2\left(1-F\right)~.
\label{eq:cnprobs}
\end{equation}
Then, the aggregate probability of shift for a given lottery pair is 
\begin{equation}
\mathbb{P}\left(\text{shift}\right)=
2F p_1\left(1-p_1\right)+2(1-F) p_2\left(1-p_2\right)~,
\label{ryhythgwqq}
\end{equation}
which is always smaller than its homogenised version (\ref{eq:P_shift_homo}). 
This results from the concavity of the function $f(p) = p(1-p)$. In the case 
$F=1/2$, this is also straightforwardly seen from the inequality 
$\left(p_1^2+p_2^2\right)/2\geq p_1p_2$. The equality between expression 
(\ref{ryhythgwqq}) and (\ref{eq:P_shift_homo}) with (\ref{eq:cnprobs}) is 
recovered obviously for the homogeneous case, i.e. for $F=0$ or $F=1$, or 
for $p_1=p_2$.

We now propose a simple quantitative model by assuming the following ansatz for 
$p_1$ and $p_2$:
\begin{equation}
\left\{
\begin{aligned}
p_1 &= p + \alpha p\left(1-p\right) & \;\;\alpha\in\left[0,1\right]\\
p_2 &=p - \beta p\left(1-p\right) & \;\;\beta=\alpha F /\left(1-F\right)\in\left[0,2\right]
\end{aligned}\right.
\label{eq:P_shift_hetero}
\end{equation}
where the value for $\beta$ derives from (\ref{eq:cnprobs}). Intuitively, the 
ansatz $p_1= p + \alpha p\left(1-p\right)$ in (\ref{eq:P_shift_hetero}) states that 
the first group of decision makers tends to follow and overweight the majority choice 
when the two lotteries are difficult to tell apart (region of $p$ not too much larger 
than $1/2$). We can refer to this first group as ``majoritarian''. The second ansatz 
$p_2 =p - \beta p\left(1-p\right)$ in (\ref{eq:P_shift_hetero}) states that the second 
group of decision makers tends to dislike the average preferred choice, the more 
difficult it is to decide between two lotteries. We call this second group ``contrarian''. 

\begin{figure*}[ht]
    \centering
    \includegraphics[width=0.8\textwidth]{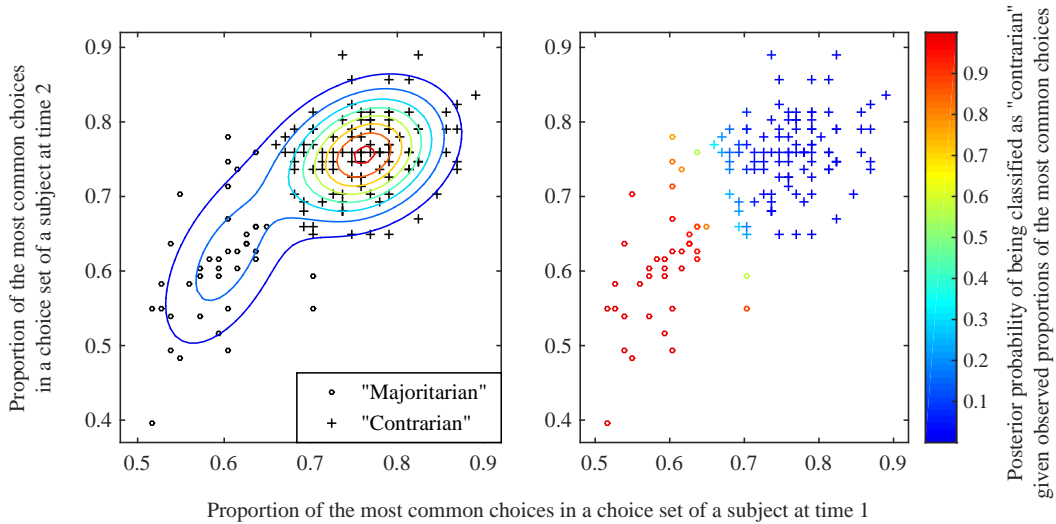}
\caption{
\small
Proportion of the most common (i.e. majority) choices in a choice set of a subject, 
observed at two repetitions of the experiment (times 1 and 2). Each of the 142 data 
points represents one decision maker. The likelihood ratio (Wilks) test rejects 
the hypothesis of a homogeneous population ($H_0$: a bivariate Gaussian model) with 
p-value $=1.4\times10^{-7}$ in favor of a heterogeneous model ($H_1$: a bivariate 
Gaussian Mixture model with 2 components). Probability density function of the latter 
is illustrated by the contour plot ({\bf left}). The bigger (resp., smaller) component 
is characterized by a higher (resp., lower) proportion of the most common choices, 
with average value over times 1 and 2 equal to 0.76 (resp., 0.61). This empirical 
feature supports assumptions of the heterogeneous model (\ref{eq:P_shift_hetero}) 
with two groups of decision makers: ``majoritarian'' (plus sign) and ``contrarian'' 
(circle). {\bf Right}: Prevailing extreme values of posterior probabilities (either 
close to 1, or to 0 of the heat map) reflect low uncertainty of assessing an observed 
decision maker to a particular group: ``majoritarian'' (103 subjects $=73\%$) and 
``contrarian'' (39 subjects $=27\%$).
}
\label{fig4}
\end{figure*}

\begin{figure*}[ht]
    \centering
    \includegraphics[width=0.5\textwidth]{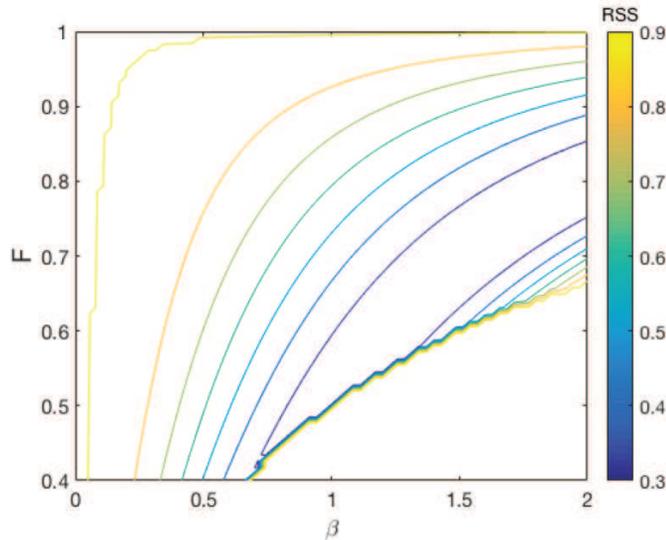}
\caption{\small 
Contour plot of the residual sum of squares (RSS) obtained during calibration of the 
heterogeneous model (\ref{eq:P_shift_hetero}). The starting values of parameters $\bt$ 
and $F$ are chosen with an iterated tabu search. The minimum $RSS_{min}=0.2331$ can be 
achieved by different combinations of the parameters. Given the estimated size of the 
``majoritarian'' group, i.e. $F=0.73$ (Figure \ref{fig4}), the optimal value of 
$\beta=1.6$.
}
\label{fig5}
\end{figure*}

Parameter $\alpha$ thus quantifies the difference between the majoritarians and 
contrarians in their tendencies to reproduce at time 2 their earlier choice at time 1. 
We do not claim that this parameterisation (\ref{eq:P_shift_hetero}) is unique or has 
a strong theoretical basis. It is offered as a simple generalization, with one additional 
parameter, to the most parsimonious model (\ref{eq:P_shift_homo}). However, as Figures 
\ref{fig4}, \ref{fig5} and \ref{fig7} show, 
this simple ansatz provides an excellent fit to the data. We stress that the determination 
of the corresponding best $\alpha$ (and thus $\beta$) and $F$ is aided by the use of the 
bivariate Gaussian mixture model shown in Figure \ref{fig4}.

First, we compare this heterogeneous model (\ref{eq:P_shift_hetero}) with the 
data by analysing decision makers with respect to their propensity to follow (or 
to oppose) the majority choice. For each decision maker, 
Figure \ref{fig4} (left subplot) shows proportion of the most 
common (i.e. majority) choices in a choice set of a subject, observed during the 
two repetitions of the experiment (times 1 and 2). In other words, for each subject 
we plot the proportion of the lottery pairs, for which the choice of a subject 
coincides with the majority choice (at time 1 and 2). For this dataset, according 
to the likelihood ratio test (the Wilks test) \cite{Wilks38}, the hypothesis of a 
homogeneous population (a bivariate Gaussian model, $H_0$) is rejected with p-value 
$=1.4\times10^{-7}$ in favor of a heterogeneous model (a bivariate Gaussian Mixture 
model, $H_1$). Probability density function of the latter is illustrated by the 
contour plot. The Gaussian mixture model has two components. The bigger (resp., 
smaller) component is characterized by a higher (resp., lower) proportion of the 
individual choices that coincide with the majority choice, with average value over 
times 1 and 2 equal to 0.76 (resp., 0.61). This feature of empirical clustering 
supports the suggested heterogeneous model (\ref{eq:P_shift_hetero}) with two 
groups of decision makers: ``majoritarian'' (plus sign) and ``contrarian'' (circle).

For the same data, Figure \ref{fig4} (right subplot) highlights 
posterior probabilities of a Gaussian mixture component (``contrarian'') given 
each observation. Prevailing extreme values of posterior probabilities (either 
close to 1, or to 0) reflect low uncertainty of assessing an observed decision 
maker to a particular group (i.e. unambiguous clustering), where 103 subjects 
($73\%$) are classified as ``majoritarian'' and 39 subjects ($27\%$) -- as 
``contrarian''.

Thus, the experimental data support the proposed classification of decision makers 
in two groups according to their propensity to follow (or to oppose) the most common 
choice (\ref{eq:P_shift_hetero}), and the estimated size of the ``majoritarian'' 
group $F=0.73$.

At the second step, the model with heterogeneity (\ref{eq:P_shift_hetero}) is 
calibrated to the same data as its homogeneous predecessor (\ref{eq:P_shift_homo}), 
which was shown in Figure \ref{fig3}. The starting values of parameters $\bt$ 
and $F$ are chosen with an iterated tabu search \cite{Glover93}. 
Recall that tabu search uses a local search procedure to iteratively move from one 
potential solution to an improved solution in the neighborhood of the starting point, 
until some stopping criterion has been satisfied. The term ``iterated'' refers to the 
fact that we start from many random initial conditions in the space of parameters. 

Optimization results for the heterogeneous model are presented in Figure \ref{fig5}. 
This contour plot illustrates that the minimum residual sum of squares 
($RSS_{min}=0.2331$) can be achieved by different combinations of the parameters 
$\bt$ and $F$. However taking into account the size of the ``majoritarian'' group 
that was estimated at the first step, i.e. $F=0.73$, the optimal value of $\bt=1.6$. 
Then, given (\ref{eq:P_shift_hetero}), $\alpha=0.6$.

\begin{figure*}[ht]
\centerline{
\includegraphics[width=0.6\textwidth]{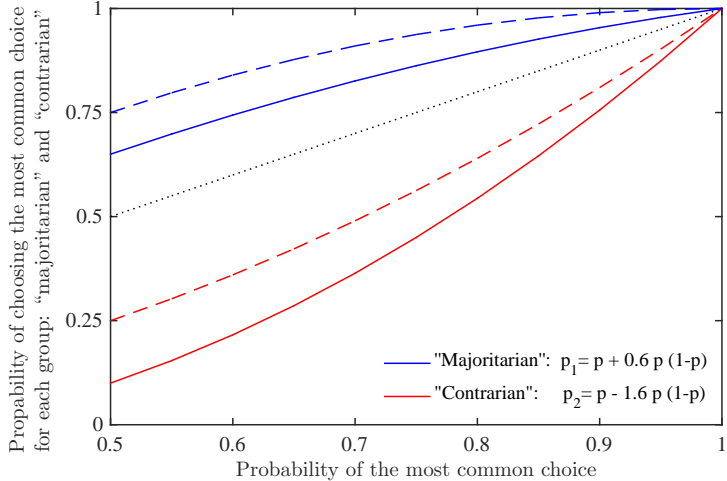}
}
\caption{\small 
Probabilities $p_1$ and $p_2$ obtained by calibrating the heterogeneous model 
(\ref{eq:P_shift_hetero}) with which the most common choice is chosen by each of the 
postulated two groups of decision makers as a function of the frequency $p$ of the 
majority choice aggregated over the whole population. The solid top (resp. bottom) 
curve shows $p_1$ (resp. $p_2$) of the ``majoritarian'' (resp. ``contrarian'') 
decision makers, with the estimated size of the ``majoritarian'' group $F=0.73$. For 
reference: dotted line is the identity line ($y=x$); dashed lines represent the case 
of equal-sized groups ($F=0.5$ leading to the best estimates $\alpha=\beta=1$).
}
\label{fig6}
\end{figure*}

\begin{figure*}[ht]
\centerline{
\includegraphics[width=0.6\textwidth]{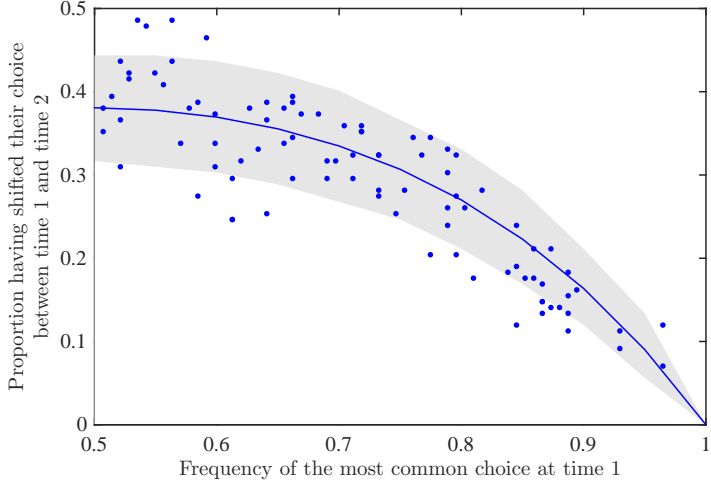}
}
\caption{\small 
Same data as in Figure \ref{fig3}, which is compared with the prediction 
(\ref{ryhythgwqq}) of a heterogeneous population of two groups of decision 
makers with $p_1$ (``majoritarian'') and $p_2$ (``contrarian'') given by 
(\ref{eq:P_shift_hetero_calib}). The proportion of group sizes is approximately 
3:1 in favor of ``majoritarian'', i.e. estimated $F=0.73$. The shaded area 
represents the 5\% and 95\% quantiles, i.e. the area where 90\% of the shifts 
should fall according to Monte Carlo simulations using the above heterogeneous 
model (3000 simulations per pair of lotteries).
}
\label{fig7}
\end{figure*}

After its calibration, the model with heterogeneity (\ref{eq:P_shift_hetero}) is 
expressed as
\begin{equation}
\left\{
\begin{aligned}
p_1 &=p + 0.6 p\left(1-p\right)\\
p_2 &=p - 1.6 p\left(1-p\right) ~,
\end{aligned} \right.
\label{eq:P_shift_hetero_calib}
\end{equation}
which is represented in Figure \ref{fig6}. As intended, the decision makers 
referred to as ``majoritarian'' tend to follow the most common choice (average 
value of $p_1\approx 0.85$). In contrast, the decision makers that we call 
``contrarian'' tend to weaken or even oppose the most common choice (average value 
of $p_2\approx 0.49$). Then, given equation (\ref{eq:P_shift_homo}), the average 
probability of shift for ``majoritarian'' group ($\approx0.26$) in much lower, than 
for the ``contrarian'' group ($\approx0.50$).

Finally, Figure \ref{fig7} presents the same data as Figure \ref{fig3} but 
now the model (\ref{eq:P_shift_hetero}) is taking into account the heterogeneity of 
population, differentiating between the two groups of decision makers: ``majoritarian'' 
($\approx 3/4$) and ``contrarian'' ($\approx 1/4$), with, respectively, $p_1$ and 
$p_2$ given by (\ref{eq:P_shift_hetero_calib}). The grey band represents the 90\% 
confidence interval, which is delineated by the 5\% and 95\% quantiles, i.e. the area 
where 90\% of the shifts should fall according to Monte Carlo simulations using the 
above model with two groups (3000 simulations per pair of lotteries). This allows us 
to quantify the uncertainty band resulting from sampling variabilities at times 1 and 
2, using standard Bernouilli statistics. While this model is clearly over-simplified, 
it provides an excellent fit to the data confirming that, within the probabilistic 
choice framework, heterogeneity among decision makers is sufficient to account 
quantitatively for the observed changes of behaviour between repetitions of the 
experiment (times 1 and 2). 

\section{Calibration of quantum decision theory}\label{sec:QDT}

\subsection{Brief presentation of stochastic cumulative prospect theory (logit-CPT) 
and quantum decision theory (QDT)}

Based on the analysis of choice reversals in a repeated experiment, the previous 
section has shown that the hypothesis that decisions are probabilistic provides a 
parsimonious and quantitative description of decision making. We thus endeavour to 
test two probabilistic choice theories, (i) stochastic cumulative prospect theory 
(logit-CPT) and (ii) quantum decision theory. Both theories are summarised in the 
Appendix. 

Here ``quantum'' model extends the ``classical'' CPT by including an attraction 
factor, which accounts for interfering choice options and a state of mind. These 
``quantum'' interference effects explain with a single origin the observed 
systematic deviation from predictions of classical models. Nested models allow 
for straightforward quantitative comparison.

Prospect theory \cite{KaTversky,KaTversky2} is now the most famous alternative to 
expected utility theory. The outcomes are quantified through a value function $v$, 
weighted by subjective probabilities obtained from the objective probability via 
a non-additive weighting function $w$. Moreover, the value function separates gains 
and losses, where the notions of gains and losses are defined with respect to a 
reference point, here assumed to be zero. Cumulative prospect theory (CPT) can be 
combined with a probabilistic choice function, allowing for probabilistic deviations 
from the option that maximises the choice criterion with respect to alternative 
options. There are many probabilistic extensions of CPT, some of which are modeling 
something entirely separate from response errors using polyhedral combinatorics, 
such as, e.g. in \cite{Regenwetteretal14}. The probabilistic version of CPT that 
we use here is called logit-CPT because the probability weighting scheme uses the 
logit function (see Appendix and below). Such stochastic extension is often 
perceived as an add-on to an intrinsically deterministic CPT approach that is 
necessary to account for the observed stochasticity of human choices, interpreted 
as errors or unobserved components of an underlying deterministic process.

Quantum decision theory (QDT) is based on two essential ideas: (a) an intrinsic 
probabilistic nature of decision making and (b) a generalisation of probabilities 
using the mathematics of Hilbert spaces that naturally account for entanglement 
between choices  \cite{YSQDT08,YSentropy,YSmathQDT,YSPosOp}. Thus, in contrast 
to logit-CPT, it places the probabilistic nature of choice at the center of its 
construction. As recalled in the Appendix (see expressions (\ref{eq:QDTsum1}-
\ref{eq:QDTsum})), a fundamental result of QDT is that the probability 
$p\left(\pi_n\right)$ of a given prospect $\pi_n$ can in general be decomposed 
as the sum of two terms according to
\be
p\left(\pi_n\right)=f\left(\pi_n\right)+q\left(\pi_n\right)~.
\label{nhtnbwg}
\ee
The first term $f(\pi_n)$ is associated with the utility of the prospect under 
consideration and, therefore, is called the {\it utility factor}. The second 
term $q(\pi_n)$ accounts for interference and entanglement between prospect and 
state of mind, and results technically from the complex quantum nature of the 
probabilities describing the choices of decision makers. In decision theory, it 
characterizes subjective and subconscious processes of the decision maker related 
to other available prospects, as well as past experiences, beliefs and momentary 
influences, and is referred to as the {\it attraction factor}. We interpret the 
attraction factor as representing a subconscious attraction of a person to a given 
prospect. The attraction depends on the state of mind that can be influenced by 
external (i.e. situational) and/or internal (i.e. hunger, mood, fatigue, etc.) 
factors. For more precise definitions of the attraction factor, we refer to the 
Appendix and to \cite{YSQDT08,YSentropy,YSmathQDT,YSPosOp}.

By the quantum-classical correspondence principle, when the quantum term $q(\pi_n)$
becomes zero, the quantum probability reduces to the classical probability, so that 
$p(\pi_n)\to f(\pi_n)$ for $q(\pi_n)\to 0$, with the normalization $\sum_n f(\pi_n)=1$, 
with $0 \leq f(\pi_n) \leq 1$. In the sequel, we use a logit-CPT form for the utility 
factor $f(\pi_n)$ given by expression (\ref{tjytwrg}) below, which corresponds to 
the first term in equation (\ref{netyhnhwgrqg1}). We assume that logit-CPT can 
adequately characterize the utility of an isolated prospect for a decision maker. 
While logit-CPT incorporates some subjective deviations of values and probabilities, 
it treats each prospect separately, with no interference between the different 
prospects or no interference between a given prospect and the state of mind.

In QDT model, these interdependencies are incorporated via the attraction factor, 
which embodies the additional complex unconscious deliberations and preferences 
associated with decision making. By construction, it enjoys the following properties 
\cite{YSQDT08,YSentropy,YSmathQDT,YSPosOp}. It lies in the range 
$-1\leq q(\pi_n)\leq 1$ and satisfies the {\it alternation law} $\sum_n q(\pi_n)=0$. 
In addition, for a large class of distributions, there exists the {\it quarter law}
\be
\frac{1}{N} \sum_{n=1}^N | q(\pi_n) | = \frac{1}{4} \;  .
\label{eq:QL}
\ee
In the presence of two competing prospects, one can show that, in the absence of any
other information (the so-called ``non-informative prior''), one obtains
\be
|q(\pi_n)| \approx 0.25~,
\label{heytnthw}
\ee 
which makes it possible to give quantitative predictions in absence of additional 
information \cite{YSQDT1ThDec11,YSQDT14cogn,YSQDT15Philtrans,YSQDT15frontpsy}. In 
the following, we go beyond (\ref{heytnthw}) and introduce a mathematical expression 
(\ref{eq:qfac}) with constant absolute risk aversion (CARA) utility function 
(\ref{eq:CARA100}) for the attraction factor, which corresponds to the second term 
in equation (\ref{netyhnhwgrqg2}) and is motivated by the structure of the pairs of 
lotteries presented to the decision makers.

\subsection{Methodology to estimate logit-CPT and QDT}

We follow and extend the procedure of parameters estimation proposed by 
\cite{HMLMurphy}. We first summarise their method and then extend it to QDT.
Here, we use the same data set as studied in \cite{HMLMurphy} to allow for a precise 
comparison and thus evaluation of the possible gains provided by quantum decision 
theory. The proposed QDT parameterization is obviously applicable to other data sets 
and we encourage readers to apply it to their own data sets.
		
According to stochastic decision theories, such as logit-CPT and QDT, the option 
$A_j$ of the pair $j$ of lotteries is chosen by a subject over the option $B_j$  
with a probability $p_{A_j}$, which depends on individual parameters. These 
parameters can be estimated by fitting the model to the data obtained at time 1 
and then used for predicting the outcomes at time 2.

The answers from the decision maker $i\in\left\{1\dots142\right\}$ at time 1 are 
denoted $\left(\Phi_j^i\right)_{j=1}^{91}$:
\begin{equation}
\Phi_j^i=\left\{
\begin{aligned}
0\text{ if subject $i$ chooses $A$ in the $j^\text{th}$ gamble},\\
1\text{ if subject $i$ chooses $B$ in the $j^\text{th}$ gamble}.
\end{aligned} \right.
\end{equation}

Given the choices $\left(\Phi_j^i\right)_{j=1}^{91}$, the individual parameters 
of the decision maker $i$ can be estimated with a maximum likelihood method. A 
natural choice for the objective function is
\begin{equation}
\Pi^i=\prod_{j=1}^{91} p_{A_j}^{1-\Phi_j^i}p_{B_j}^{\Phi_j^i}~.
\end{equation}
However, it has been shown \cite{NilssonRiesWagen11} that this optimization 
method gives unreliable estimates at the individual level, since a shift of a 
single answer sometimes leads to very different parameters estimates. The 
\emph{hierarchical maximum likelihood} method based on the work of \cite{Farrell} 
fixes this issue by introducing the assumption that the individual parameters are 
distributed in the population with a given density distribution. The optimization 
is then performed for each subject, weighting the objective functions with the 
density distributions obtained at the population level. In Ref. \cite{HMLMurphy}, 
this method has been applied to the experimental data described in section 
\ref{subsec:expdesign}. Applied to stochastic CPT briefly described in the Appendix, 
the distributions of the parameters $\alpha$, $\lambda$, $\gamma$ and $\delta$ were 
assumed to be lognormal. Each log-normal distribution is defined through its location 
parameter $\mu$ and its scale parameter $\sigma$, which were estimated with a maximum 
likelihood method at the aggregate level.

The exactly same data and parameters estimation procedure were used in the 
analysis of the present article, which allows for a direct comparison of 
stochastic cumulative prospect theory and quantum decision theory. For 
stochastic cumulative prospect theory, we are able to recover precisely the 
quantitative results reported by \cite{HMLMurphy}. In other words, we did not 
use the parameters reported by \cite{HMLMurphy} but re-estimated them ourselves 
completely independently, reproducing entirely the whole calibration procedure 
for the logit-CPT. Then, we extended the procedure to calibrate and test QDT 
as explained below. The detailed description of the methodology follows.

\paragraph{$\bullet$ At the aggregate level}
~~\\
At the aggregate level, the parameters are estimated with a maximum likelihood 
method for both models (logit-CPT and QDT). The objective function is 
\begin{equation}
\Pi^{\text{agg}}=
\prod_{i=1}^{142}\prod_{j=1}^{91} p_{A_j}^{1-\Phi_j^i}p_{B_j}^{\Phi_j^i}~,
\end{equation}
where the probability of choosing option $A$ over option $B$ is defined as follows 
(see Appendix):

\begin{itemize}
\item
-- for logit-CPT: 
\be
p_{A_j}=\frac{1}{1+e^{\varphi\left(\tilde U_{B_j}-\tilde U_{A_j}\right)}}~,
\label{tjytwrg}
\ee
\item
-- for QDT:
\be 
p_{A_j}=
\frac{1}{1+e^{\varphi\left(\tilde U_{B_j}-\tilde U_{A_j}\right)}}+
\min \left(f_{A_j},1-f_{A_j}\right)\tanh\left(a\left(U_{A_j}-U_{B_j}\right)
\right)~.
\label{netyhnhwgrqg1}
\ee
\end{itemize}

To be clear, associated with the utility factor, ${\tilde U}$ represents the 
utility according to the CPT framework defined by expression (\ref{eq:utiPT}), 
while $U$, which is defined by expression (\ref{eq:CARA100}) as the CARA function 
with a coefficient of absolute risk aversion $\eta$, enters into the definition 
(\ref{eq:qfac}) of the attraction factor.

Note that the QDT formulation has two additional parameters ($a$ and $\eta$) 
compared to logit-CPT, so that the later is nested in QDT (it is retrieved from 
the QDT formulation by setting $a=0$).

\paragraph{$\bullet$ At the individual level}
~~\\
When applied finally to the individual level, the parameters are estimated with 
a hierarchical maximum likelihood method for both models (logit-CPT and QDT). In 
a nutshell, this means first estimating the distribution of parameters at the 
aggregate level to obtain prior distributions, which are then used as weights 
penalising possible over-determinations at the individual level. The objective 
function for each subject $i$ is
\begin{equation}
\Pi^i=
g_\alpha g_\lambda g_\gamma g_\delta
\prod_{j=0}^{91} p_{A_j}^{1-\Phi_j^i}p_{B_j}^{\Phi_j^i}~,
\end{equation}
where $g_X$ is the distribution of the parameter $X\in\{\al,\lbd,\gm,\dlt\}$, 
according to the experimental results from \cite{HMLMurphy}. The probability of 
choosing option A over option B is defined as follows:

\begin{itemize}
\item
-- for logit-CPT identically as for the aggregate level in (\ref{tjytwrg}): 
\be
p_{A_j}=\frac{1}{1+e^{\varphi\left(\tilde U_{B_j}-\tilde U_{A_j}\right)}}~,
\ee
\item
-- for QDT: 
\be 
p_{A_j}=
\frac{1}{1+e^{\varphi(\tilde U_{B_j}-\tilde U_{A_j})}}+
\min \left(f_{A_j},1-f_{A_j} \right) 
\tanh \left( a^{\text{agg}} \left( U_{A_j}^{\text{agg}}-
U_{B_j}^{\text{agg}} \right) \right),
\label{netyhnhwgrqg23d}
\ee
\be 
p_{A_j}=\frac{1}{1+e^{\varphi\left(\tilde U_{B_j}-\tilde U_{A_j}\right)}}+
\min \left( f_{A_j},1-f_{A_j} \right) \tanh \left( a\left(U_{A_j}-U_{B_j} \right)
\right)~,
\label{netyhnhwgrqg2}
\ee
where the exponent ``$\text{agg}$" indicates that, at the individual level, $a$ and 
$\eta$ are not seen as parameters, but replaced by their optimal values found at the 
aggregate level.
\end{itemize}

In particular, at the individual level, the QDT formulation involves the same number 
of individual parameters as the logit-CPT formulation.

The solver used for all the optimizations is the \textit{fminsearch} function from 
MATLAB (Nelder \& Mead simplex algorithm), the starting values of the parameters are 
chosen with a tabu search.

\subsection{Calibration and prediction at the aggregate level}

At the aggregate level, the optimization problem for QDT involves seven 
parameters: five for the QDT utility factor, equation (\ref{eq:ffac}), which is 
identical to the logit-CPT formulation, and two original parameters for the QDT 
attraction factor, equation (\ref{eq:qfac}). Thus, the logit-CPT model is nested 
in the QDT one (null hypothesis: $a^{\text{agg}}=0$) \cite{GourierouxMonfort94}. 
This implies that one has to be very careful with choosing a statistical test, so 
that it can ``punish'' the more general formulation (i.e., the unrestricted 
QDT model). Widely used methods of a relative quality estimation for the models 
with different number of parameters include Akaike information criterion (AIC) 
and Bayesian information criterion (BIC). Both criteria compare models' goodness 
of fit, which is assessed by the likelihood function, while penalising for a larger 
number of estimated parameters. However for nested hypothesis, the AIC and BIC are 
superseded by the likelihood ratio test, which is in fact the most powerful test 
among competitors. We apply the log-likelihood ratio test (the Wilks test) to 
compare QDT and logit-CPT models. For nested hypotheses, one can show that two 
times the log-likelihood ratio has a chi-square distribution with a number of 
degrees of freedom equal to the difference in the number of parameters between 
QDT and logit-CPT (which is $2$, $a$  (\ref{netyhnhwgrqg1}) and $\eta$ (\ref{eq:CARA100})), under the null that the 
generating process is the logit-CPT (i.e., the restricted model with the smaller 
number of parameters).

Performing the log-likelihood ratio test, we find that the null hypothesis 
(logit-CPT) is rejected at the 95\% level. In other words, the logit-CPT model 
is insufficient to describe the data, and the QDT formulation with the novel 
attraction factor provides a significant improvement, which is sufficiently large 
to compensate for the ``cost'' of two additional parameters.

\begin{figure*}[ht]
\centerline{
\includegraphics[width=0.7\textwidth]{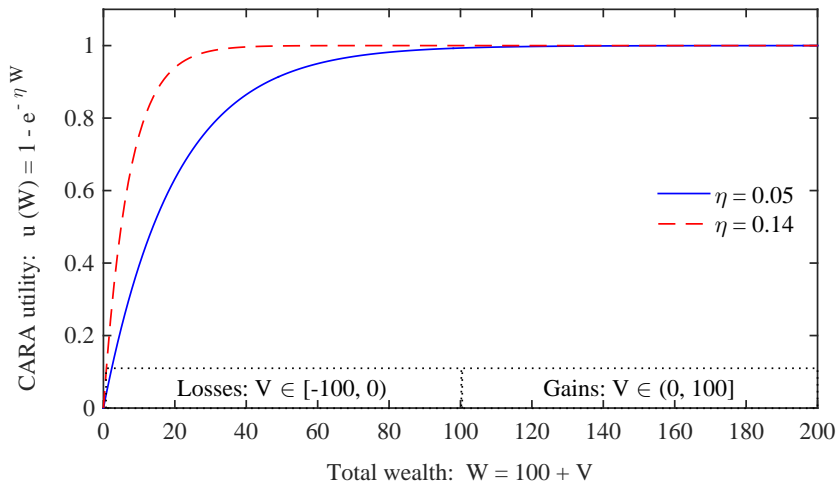}
}
\caption{\small 
Constant absolute risk aversion (CARA) utility function. In the experiment, the 
outcomes $V$ of choice options (i.e. binary lotteries defined in table \ref{tab:gamble}) 
are between $-100$ and $100$, and the initial endowment $W_0=100$, thus, the total 
wealth $W\in[0,200]$. With this utility function and expression (\ref{eq:qfac}), with 
$\eta=0.05$ (for the majority of decision makers) the attraction factor $q$ is small 
except for pairs of lotteries involving big losses. Higher value of $\eta=0.14$ (for 
``contrarian'' group) reduces this $q$ effect to a narrow range of extreme losses.
}
\label{fig8}
\end{figure*}

For the two models, the values of the parameters estimated at the aggregate level 
for all participants (i.e. with the assumption of homogeneity) are highlighted in 
bold in Table \ref{tab:Calibr-param-All-Hetero}. For the QDT attraction factor $q$, 
the CARA utility function $U$ with the obtained parameter $\eta=0.05$ is illustrated 
in Figure \ref{fig8}. In particular, since $|q|$ depends on the utility 
difference $U_A-U_ B$ of the alternative lottery options $A$ and $B$, the attraction 
factor is small except for some gambles involving big losses. Thus, the QDT 
attraction factor accounts for the experimental observation that decision makers 
do not care much about medium payments, but respond to large losses. Note that the 
typical size of the CARA coefficient $\eta$ is set by the size of the initial wealth 
and of the payoffs of the lotteries. 

Most of the parameters describing the QDT utility term ($\al$, $\gm$, $\dlt$ and 
$\varphi$) are close to those obtained with logit-CPT. However, for QDT the loss 
aversion parameter $\lambda$ is smaller. This means that within QDT, though losses 
loom larger than gains in general ($\lambda>1$), a part of this effect, namely, 
aversion to big losses, is transferred to the QDT attraction factor ($q\neq 0$).

\begin{table}[ht]
\caption{\small Estimated at the aggregate level, parameters of logit-CPT and QDT. 
For ALL decision makers (in bold), values of $\al,\dlt,\gm,\varphi$ parameters are 
close for both models. The loss aversion parameter $\lbd$ is smaller with QDT (though 
$>1$) because aversion to large losses is transferred to the QDT attraction factor 
($a$ (\ref{netyhnhwgrqg1}) and $\eta$ (\ref{eq:CARA100})). Heterogeneity is introduced by classifying decision makers into 
``majoritarian'' ($\approx 3/4$) and ``contrarian'' ($\approx 1/4$) based on their 
propensity to follow/oppose the majority choice (model (\ref{eq:P_shift_hetero}), 
Figure \ref{fig4}). For ``majoritarian'' group, the QDT attraction 
factor (i.e. aversion to big losses) is even more acute, increasing the relevance 
of QDT. In contrast, for ``contrarian'' group, the QDT attraction factor has less 
impact due to the decrease in $a$, bringing closer predictions of both models.}
\label{tab:Calibr-param-All-Hetero}
\centering 
\begin{tabular}{rlccccccc}
                     & Decision makers:                     & $\alpha^{\text{agg}}$ & $\lambda^{\text{agg}}$ & $\delta^{\text{agg}}$ & $\gamma^{\text{agg}}$ & $\varphi^{\text{agg}}$ & $a^{\text{agg}}$     & $\eta^{\text{agg}}$  \\ \hline
logit-CPT            & ALL,                                 & \bf{0.73}                  & \bf{1.11}                   & \bf{0.88}                  & \bf{0.65}                  & \bf{0.30}                   & \bf{-}                    & \bf{-}                    \\
\multicolumn{1}{l}{} & including:                           & \multicolumn{1}{l}{}  & \multicolumn{1}{l}{}   & \multicolumn{1}{l}{}  & \multicolumn{1}{l}{}  & \multicolumn{1}{l}{}   & \multicolumn{1}{l}{} & \multicolumn{1}{l}{} \\  \cline{2-9} 
                     &~~~``majoritarian'' & 0.72                  & 1.13                   & 0.91                  & 0.69                  & 0.41                   & -                    & -          \\
                     &~~~``contrarian''     & 0.86                  & 0.98                   & 0.78                  & 0.40                  & 0.08                   & -                    & -             \\  \hline
QDT                  & ALL,                                 & \bf{0.69}                  & \bf{1.02 }                  & \bf{0.89}                  & \bf{0.63}                  & \bf{0.37}                   & \bf{1.47}                 & \bf{0.05}                 \\
\multicolumn{1}{l}{} & including:                           & \multicolumn{1}{l}{}  & \multicolumn{1}{l}{}   & \multicolumn{1}{l}{}  & \multicolumn{1}{l}{}  & \multicolumn{1}{l}{}   & \multicolumn{1}{l}{} & \multicolumn{1}{l}{} \\  \cline{2-9} 
                     &~~~``majoritarian'' & 0.68                  & 1.03                   & 0.92                  & 0.66                  & 0.50                   & 2.07                 & 0.05                 \\
                     &~~~``contrarian''     & 0.80                  & 0.93                   & 0.77                  & 0.40                  & 0.11                   & 0.61                 & 0.14                 \\  \hline
\end{tabular}
\end{table}

\begin{figure*}[ht]
\centerline{
\includegraphics[width=0.7\textwidth]{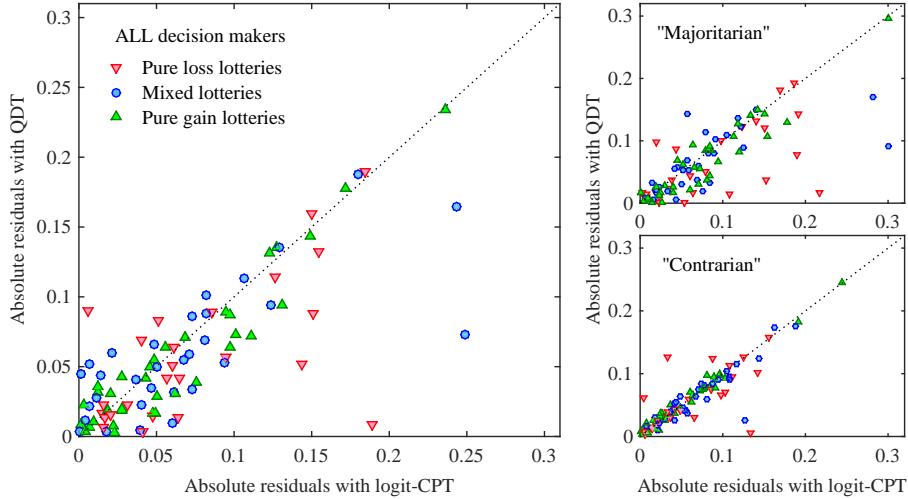}
}
\caption{\small 
{\bf Left main plot}: Distances between choice frequencies estimated at the 
aggregate level (with logit-CPT and QDT) and those observed at the first iteration 
of the experiment (time 1). Each of the 91 points is a pair of lotteries of a 
certain type: pure loss (red downward triangles), pure gain (green upward triangles) 
and mixed lotteries (blue circles). The QDT model (y-axis) performs better than 
logit-CPT (x-axis) for gambles appearing in the lower triangle below the diagonal. 
{\bf Right subplots}: The same values calculated for the two groups of decision 
makers: ``majoritarian'' and ``contrarian''. The main improvements are achieved with 
QDT for mixed and pure loss lotteries, which involve big losses, and mostly within 
the ``majoritarian'' group.
}
\label{fig9}
\end{figure*}

Figure \ref{fig9} (main plot) demonstrates performance of the two 
models regarding the types of lotteries. For each pair of gambles, absolute residuals 
of the choice frequencies, i.e. distances between their estimated values and those 
observed at the first iteration of the experiment (time $1$), are compared. For the 
QDT model absolute residuals are smaller, in particular, for mixed and pure loss 
lotteries that involve big losses. Though, for other lotteries, the improvement might 
not seem significant, table \ref{tab:rss} shows that QDT reduces the residual sum of 
squares (RSS) when summed over all gambles, as well as separately over each type of 
gambles: pure loss, pure gain and mixed lotteries. In other words, due to the fact 
that QDT discriminates between high aversion to big losses and only moderate aversion 
to small losses, QDT outperforms logit-CPT for all types of lotteries.

\begin{table}[ht]
\caption{\small For lottery types, at the aggregate level, statistics of the fit 
(time 1) and prediction (time 2) for both parametrisations -- logit-CPT and QDT. The 
QDT model outperforms in both iterations of the experiment: (i) the residual sum of 
squares (RSS) is smaller when calculated for All lottery pairs, as well as separately 
for each type -- pure loss/gain and mixed lotteries; (ii) the correlation is closer 
to 1.}
\label{tab:rss}
\centering
\begin{tabular}{@{}llrrr@{}}
Residual sum of squares (RSS) for:                &                      & logit-CPT            & QDT                  \\ \hline
ALL types of lotteries,                           & FIT (time 1)          & 0.73               & 0.52               \\
                                                  & PREDICTION (time 2)    & 0.76               & 0.59               \\
including:         &                                      & & & \\ \hline
~~~pure loss lotteries    & FIT (time 1)                        &  0.22 & 0.15                \\
                                                  & PREDICTION (time 2)          & 0.26 & 0.13                \\  \hline
~~~mixed lotteries                   & FIT (time 1)   & 0.27 & 0.17               \\
                                                 & PREDICTION (time 2)          & 0.21 & 0.18                \\  \hline
~~~pure gain lotteries       & FIT (time 1)                     & 0.24 & 0.21               \\
                                                 & PREDICTION (time 2)          & 0.29 & 0.28                \\  \hline
Correlation                                 & FIT (time 1)                         &0.93 & 0.95              \\ 
                                                 & PREDICTION  (time 2)         & 0.93 & 0.95               \\ \hline
\end{tabular}
\end{table}

\subsection{Calibration and prediction at the aggregate level for two groups of 
decision makers \label{Calib_hetero}}

Section \ref{neytbgww} demonstrated that experimentally observed choice reversals 
can be to a large degree explained by a simple probabilistic model, when heterogeneity 
of population is introduced. Decision makers were differentiated into two groups -- 
``majoritarian'' and ``contrarian'' in proportion $\approx 3:1$, -- based on their 
propensity to follow/oppose the majority choice (model (\ref{eq:P_shift_hetero}), 
Figure \ref{fig4}). In the current Section, parameters of the two 
models --- logit-CPT and QDT -- are calibrated for each group separately.

Table \ref{tab:RSS_all_hetero} compares the obtained residual sum of squares (RSS) 
of the fit (time 1) and prediction (time 2) for both models. The RSS is consistently 
lower with the QDT model for all decision makers, as well as for each group. At the 
same time, introducing heterogeneity slightly increases the RSS in comparison with 
the assumption of homogenous population, especially for predictions, which may 
indicate overfitting. The only improvement in the RSS is observed for the fit of 
``contrarian'' group with logit-CPT, so that for this group results of both models 
become quite close to each other.

\begin{table}[ht]
\caption{\small Accounting for heterogeneity, at the aggregate level, the residual 
sum of squares (RSS) of the fit (time 1) and prediction (time 2) for both 
parametrisations -- logit-CPT and QDT. For both groups, the RSS is consistently 
lower with QDT than with logit-CPT. However, the RSS slightly increases in comparison 
with the assumption of homogenous population (All decision makers), especially for 
the prediction, which indicates overfitting. The only improvement of the RSS is 
observed for the fit of ``contrarian'' group with logit-CPT, so that for this group 
results of both models become close to each other.}
\label{tab:RSS_all_hetero}
\centering
\begin{tabular}{@{}llrrr@{}}
Residual sum of squares (RSS) for:                &                      & logit-CPT            & QDT                  \\ \hline
ALL decision makers,               & FIT (time 1)                  & 0.73               & 0.52               \\
                                                 & PREDICTION (time 2)          & 0.76               & 0.59               \\
including:         &                                      & & & \\ \hline
~~~``majoritarian''                & FIT (time 1)                 & 0.96                & 0.66                \\
                                                 & PREDICTION (time 2)          & 1.03                & 0.78                \\  \hline
~~~``contrarian''                       & FIT (time 1)                 & 0.59               & 0.53               \\
                                                 & PREDICTION (time 2)         & 1.04                & 0.97                \\  \hline
\end{tabular}
\end{table}

Parameters of the logit-CPT and QDT models, estimated at the aggregate level for 
each group are presented in Table \ref{tab:Calibr-param-All-Hetero}. Within both 
models, classification of participants into two groups affected the estimates of 
the parameter $\varphi$, i.e. steepness of the logit choice function (\ref{eq:ffac}). 
``Majoritarian'' decision makers have the highest $\varphi$ that reveals a higher 
degree of conviction in their own choice. This finding is in agreement with the 
heterogeneous shift model (\ref{eq:P_shift_hetero}) and Figure \ref{fig6}, 
which predict a lower probability of choice shift between repetitions of the 
experiment for ``majoritarian'' group (Section \ref{neytbgww}). For the ``contrarian'' 
participants, the low value of $\varphi$ is in line with the prediction of a larger 
probability of the choice shift.

Within QDT model, the most noticeable distinction between the two groups concerns 
the attraction factor $q$, which captures aversion to large losses. For 
``majoritarian'' group, the QDT attraction factor is even more acute, than in the 
case of a homogenous population. Specifically, ``majoritarians'' are susceptible 
to losses of the same magnitude (the same $\eta$ (\ref{eq:CARA100})), but are more sensitive to them 
(larger $a$). In contrast, for ``contrarian'' group, the QDT attraction factor has 
small impact. Higher value of $\eta=0.14$ reduces the range of losses, to which 
participants are susceptible: ``contrarians'' are averse only to extreme losses, 
Figure \ref{fig8}. Moreover, they are less sensitive due to the decrease in 
parameter $a$. The small impact of the QDT $q$ factor on the ``contrarian'' group 
explains close predictive power of logit-CPT and QDT models. 

This distinction between the groups becomes evident on the 
Figure \ref{fig9} (right subplots). QDT visibly improves the fit 
for the ``majoritarian'' group in mixed and pure loss lotteries with big losses, 
due to the larger attraction factor. At the same time, ``contrarian'' decision 
makers are modeled similarly by both models.

\subsection{Calibration and prediction at the individual level}

\begin{figure*}[ht]
\centerline{
\includegraphics[width=0.7\textwidth]{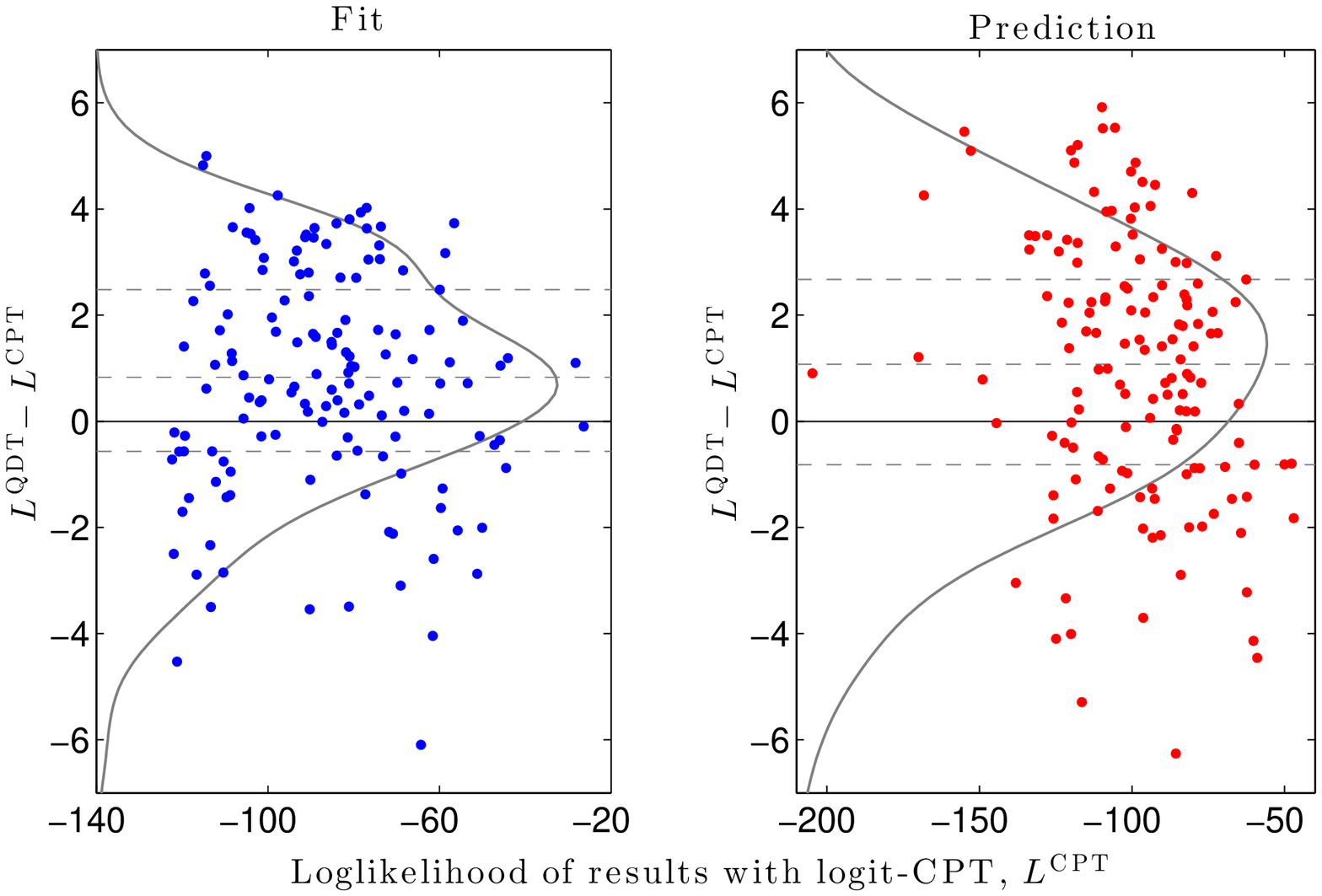}
}
\caption{\small 
Difference of log-likelihoods (y-axis, QDT minus logit-CPT) as a function 
of the log-likelihoods obtained with logit-CPT (x-axis). There are 142 
points in each plot, each point represents a decision maker. QDT performs 
better for points with positive y-coordinate (i.e. when QDT leads to a 
larger log-likelihood). The left plot shows the results of the fit (time 1), 
and the right plot shows the results of the prediction (time 2). For both 
plots, the solid curves represent the kernel estimated density of the 
difference of log-likelihoods, and the dashed lines their median and 
quartiles.
}
\label{fig10}
\end{figure*}

At the individual level, since the two formulations include the same number of 
parameters, the model selection can be done according to the (log-)likelihoods: 
the preferred model is the one that has the largest (log-)likelihood. In other 
words, due to the same number of parameters, the Bayesian information criterion 
(BIC) and the Akaike information criterion (AIC) are redundant and equivalent to 
the direct comparison of the (log-)likelihoods. According to this criterion, we 
find that the QDT model has the highest predictive power: average, over decision 
makers, log-likelihood for this model is larger (table \ref{tab:IndividCalib}).
Figure \ref{fig10} provides a comparison of the individual log-likelihoods 
obtained with logit-CPT and QDT for each subject. The figure demonstrates better 
performance of the QDT model both for the fit (at time 1) and for the prediction 
(at time 2).

\begin{table}[ht]
\caption{\small Model selection based on the calibration at the individual level. 
Average over decision makers log-likelihood is larger for QDT, thus according to 
this criterion the QDT model is preferred to logit-CPT. Since the two models have 
the same number of parameters, the AIC and BIC criteria produce the same result. 
Average fractions of choices in a choice set of a subject explained (at time 1) 
and predicted (at time 2) are also larger for the QDT model.}
\label{tab:IndividCalib}
\centering
\begin{tabular}{llcc}
Average over decision makers:      &            & logit-CPT & QDT    \\ \hline
~~~ individual log-likelihood:      & FIT (time 1)       & -86.53    & -85.77 \\
                                   & PREDICTION (time 2) & -99.31    & -98.34 \\ \hline
~~~ explained (predicted) fraction of choices: & FIT (time 1)       & 0.76      & 0.77   \\
                                   & PREDICTION (time 2) & 0.73      & 0.74   \\ \hline
\end{tabular}
\end{table}

The QDT model is also selected according to the explained fraction of choices in 
a choice set of a subject. At both iterations of the experiment -- time 1 (fit)  
and time 2 (prediction) -- the average explained (predicted) fraction of choices 
is slightly larger for QDT, than for logit-CPT (table \ref{tab:IndividCalib}). 

\begin{figure*}[ht]
\centerline{
\includegraphics[width=0.6\textwidth]{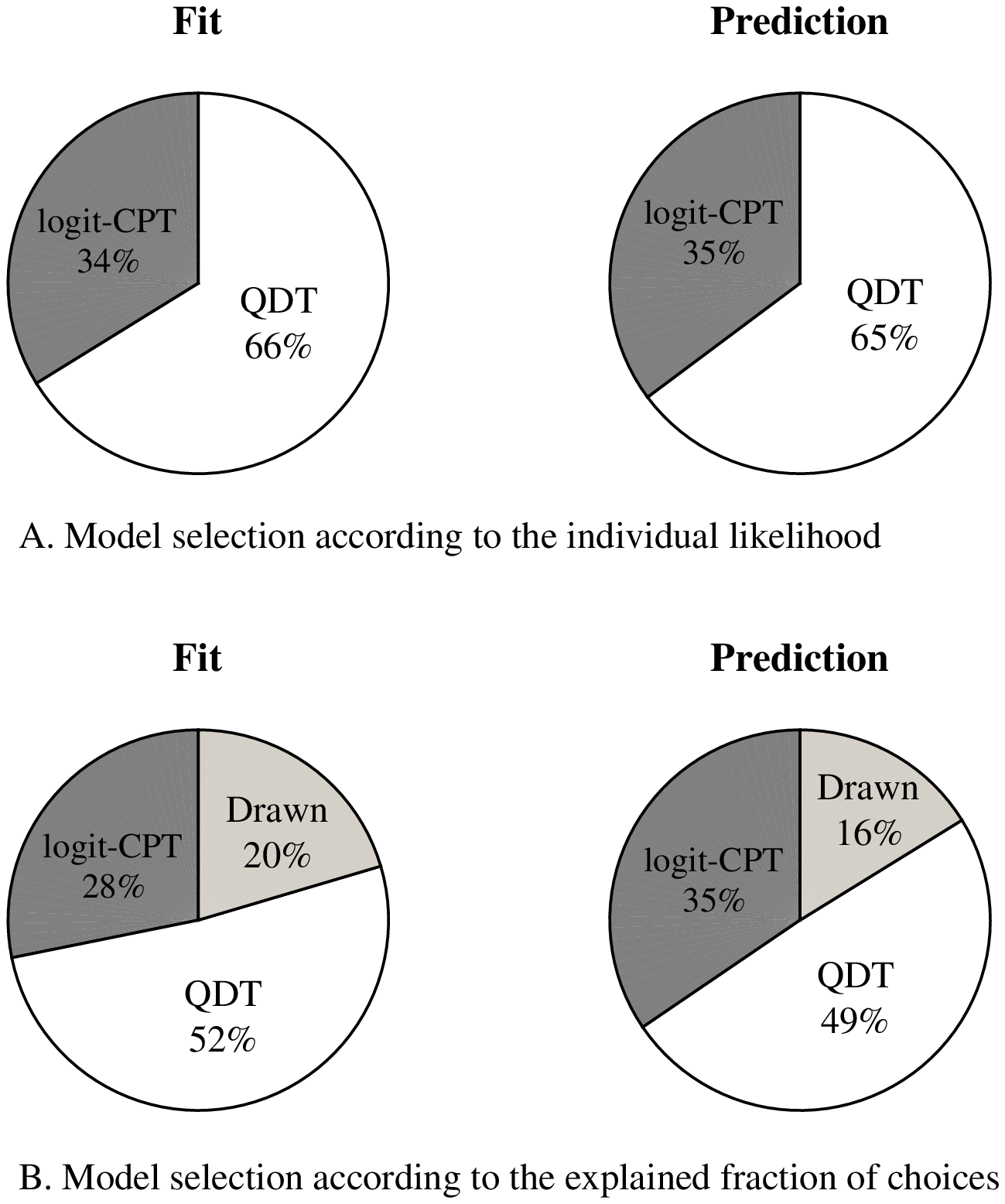}
}
\caption{\small 
On the individual basis, model selection -- QDT (white), logit-CPT (dark 
grey) or drawn among models (i.e. equal results, light grey) -- at two 
repetitions of the experiment: time 1 (fit, left) and time 2 (prediction, 
right). {\bf Upper:} According to the individual (log-)likelihood criterion, 
QDT strictly outperforms logit-CPT for $2/3$ of decision makers. {\bf Lower:} 
According to the individual explained (predicted) fraction of choices, the 
QDT model performs better for a half of subjects and at least as good as 
logit-CPT for another 20\% of participants. Thus, for both criteria the 
logit-CPT model has superior performance only for $1/3$ of the individuals.
}
\label{fig11}
\end{figure*}

Figure \ref{fig11} summarizes model selection on the individual basis. 
According to the (log-)likelihood criterion, QDT strictly outperforms logit-CPT 
for $2/3$ of decision makers. When comparing the explained (predicted) fraction 
of choices, the QDT model performs better for a half of subjects and at least as 
good as logit-CPT for another 20\% of participants. Thus, for both criteria the 
logit-CPT model has superior performance only for one third of the individuals.

\begin{figure*}[ht]
\centerline{
\includegraphics[width=0.6\textwidth]{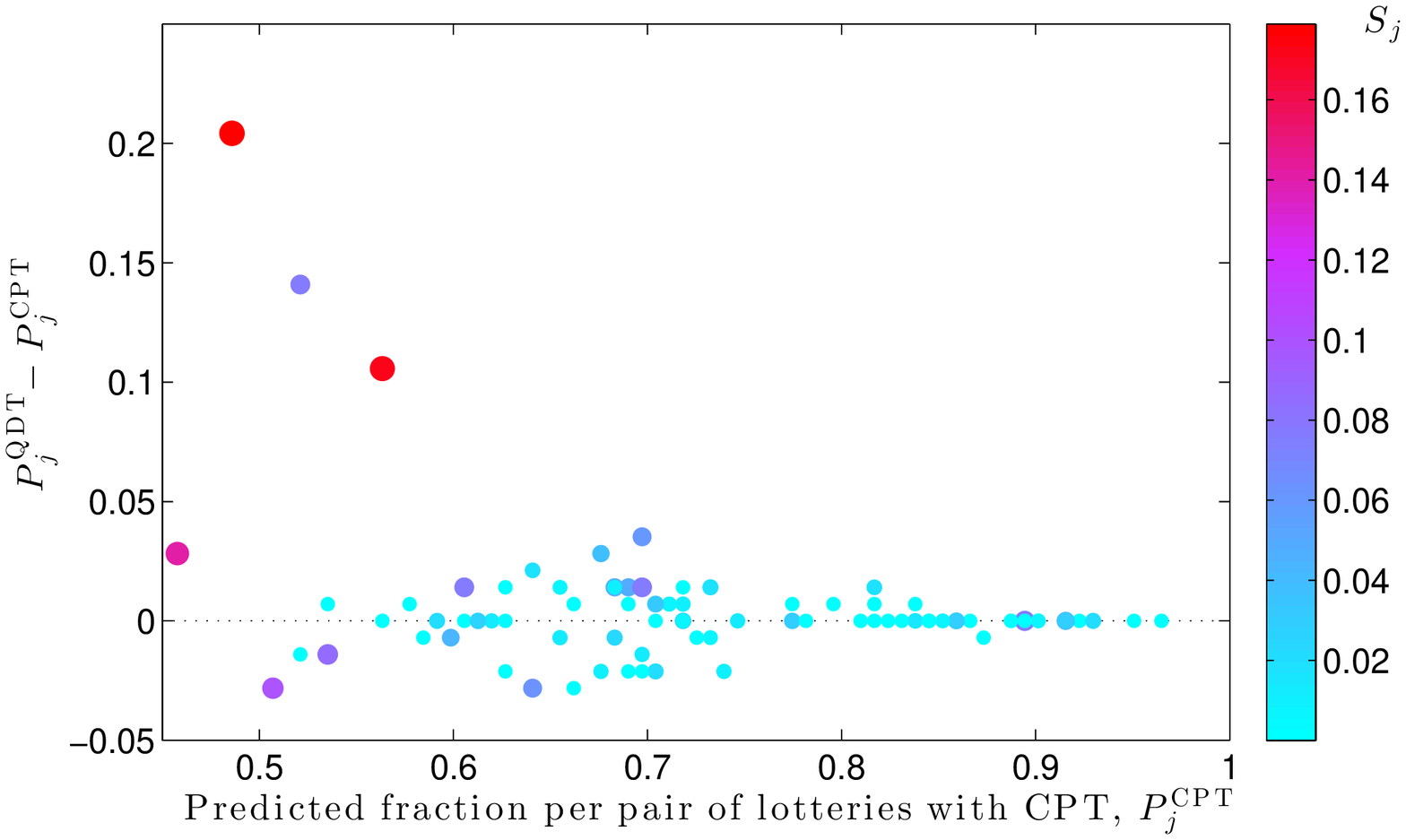}
}
\caption{\small 
$P_j^{\text{QDT}}$ (resp. $P_j^{\text{CPT}}$) represents the fraction of 
choices correctly predicted for the pair $j$ of lotteries with QDT (resp. 
logit-CPT). The difference between the two predicted fractions is plotted 
against the predicted fraction obtained with logit-CPT. For points in the 
upper part of the plot above the dotted line, QDT preforms better. The 
colors and sizes of markers encode the average intensity of the attraction 
factor among $N=142$ subjects: $S_j=\frac1N\sum_{i=1}^N |q_{A_j}^i|$.
}
\label{fig12}
\end{figure*}

\begin{figure*}[ht]
\begin{subfigure}[h]{1\textwidth}
\centerline{ 
\includegraphics[width=0.7\textwidth]{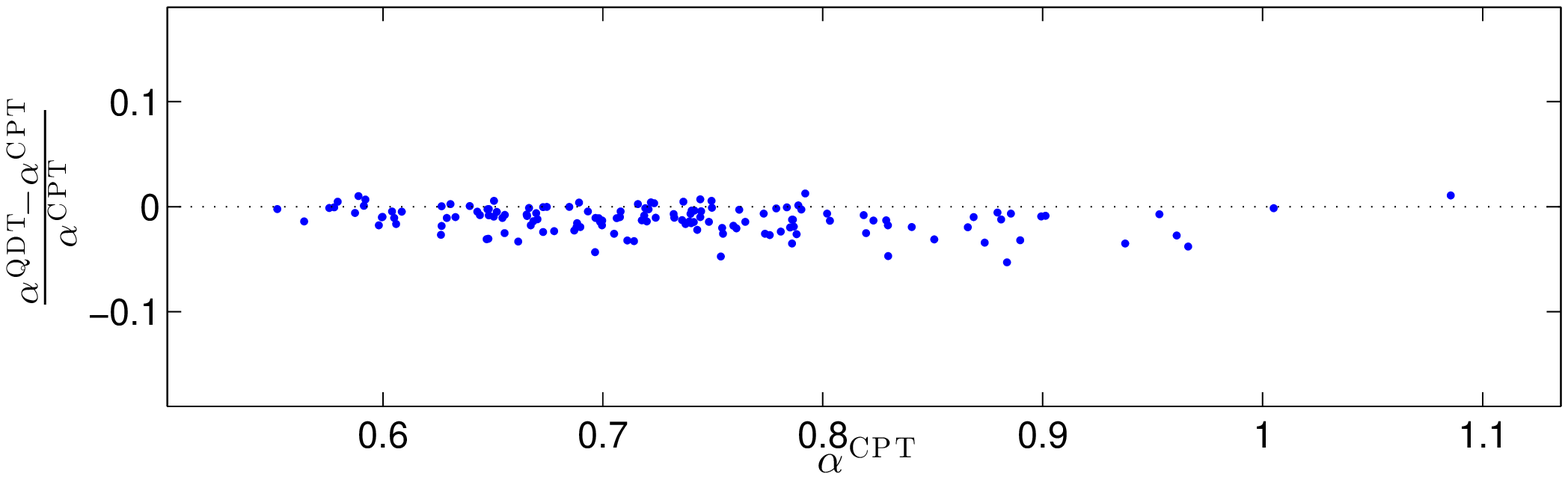} }
\end{subfigure}
\begin{subfigure}[h]{1\textwidth}\vspace{.25cm}
\centerline{ 
\includegraphics[width=0.7\textwidth]{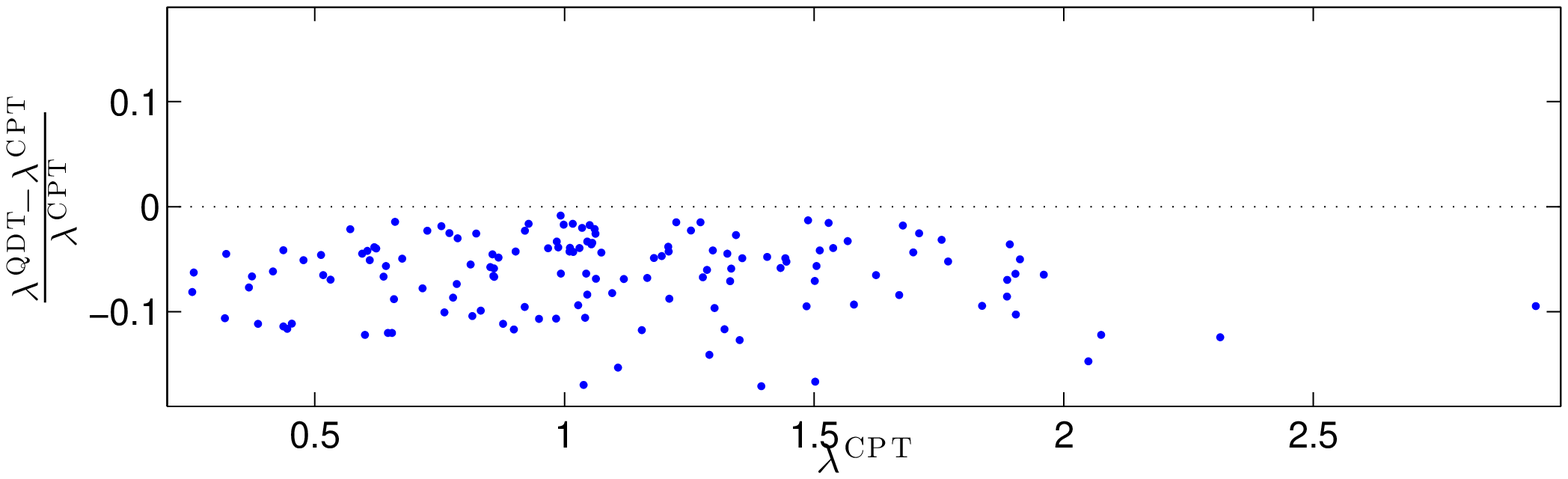} }
\end{subfigure}
\begin{subfigure}[h]{1\textwidth}\vspace{.25cm}
\centerline{ 
\includegraphics[width=0.7\textwidth]{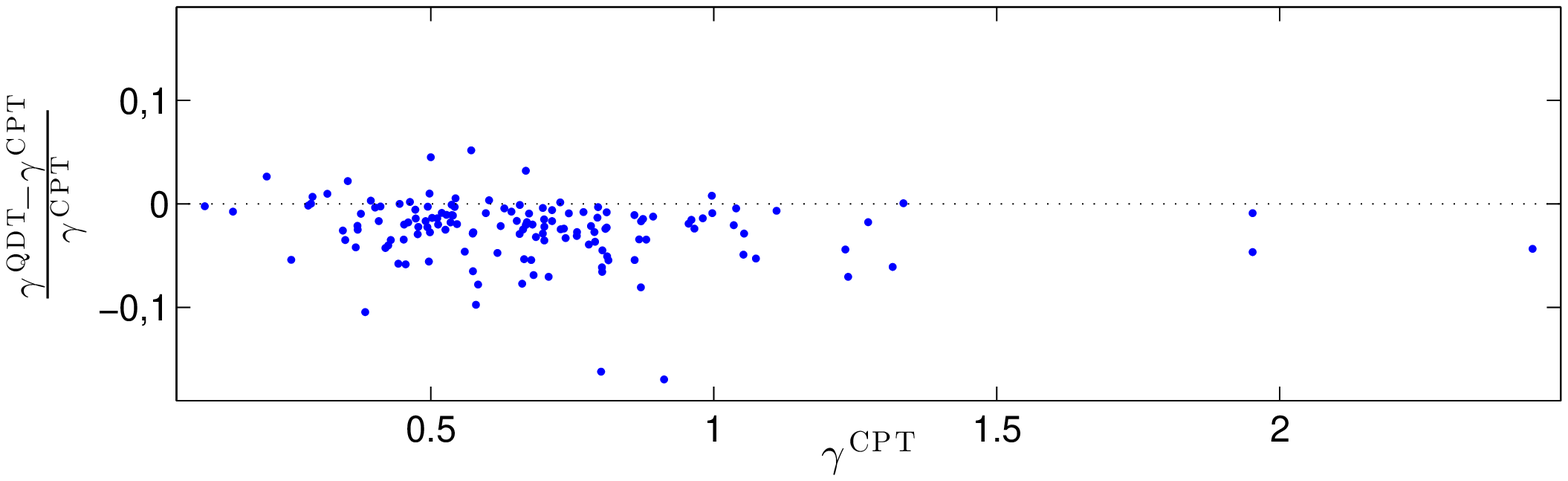} }
\end{subfigure}
\begin{subfigure}[h]{1\textwidth}\vspace{.25cm}
\centerline{ 
\includegraphics[width=0.7\textwidth]{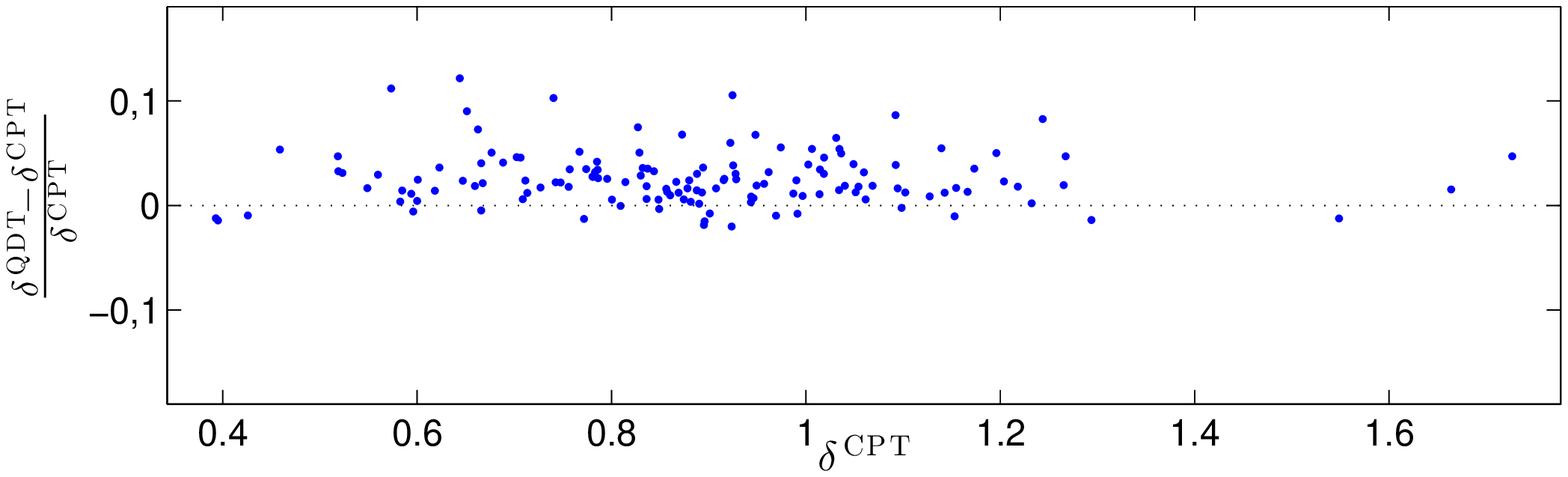} }
\end{subfigure}
\caption{\small 
Relative difference of parameters estimates (QDT minus logit-CPT) against 
estimates obtained with logit-CPT. Though for all parameters and most subjects 
the absolute relative difference are smaller than 10\%, some trends are 
noticeable. In particular, in the presence of a quantum factor for extreme 
losses, the loss aversion $\lbd$ tends to be smaller than with logit-CPT.
}
\label{fig13}
\end{figure*}

A closer look at the predicted fractions of choices for each pair $j$ of 
lotteries (Figure \ref{fig12}) reveals that the improvement obtained with 
QDT is especially noticeable in some gambles including big losses. For those 
particular gambles, the quantum attraction factor is very significant. For the 
other gambles, the predictions are of the same quality with both methods. Indeed, 
the individual parameters obtained for the QDT utility factor tend to differ by 
less than 10\% from those obtained with logit-CPT (see Figure \ref{fig13}): 
this implies that, for lotteries with negligible attraction, QDT gives individual 
predictions that are close to those given by logit-CPT.

In conclusion, at both aggregate and individual levels, the QDT model, with 
the attraction factor that captures aversion to large losses, outperforms the 
logit-CPT model according to a number of criteria. While the improvements of 
these diagnostics obtained with QDT over logit-CPT are not large, they are of 
the same size as those obtained by \cite{HMLMurphy} in their evaluation of 
different competing models (excluding QDT). In section \ref{yjnheew}, we propose 
an explanation for these results, i.e. difficulty in considerable improvement of 
choice prediction, based on an intrinsic limit of predictability associated with 
the intrinsic probabilistic nature of decision making.

\section{Limits of predictability with probabilistic choices \label{yjnheew}}

We return to the considerations and tests of Section \ref{uymjt} that strongly 
suggest that decisions are probabilistic rather than deterministic. We test further 
this hypothesis and show that it allows us to quantitatively account for the limits 
of predictability observed in the experiments.  

Indeed, Table \ref{tab:IndividCalib} and Figures \ref{fig10}-\ref{fig11} show that 
the current analytical formulation of QDT allowed us to improve the individual fit 
and prediction for most subjects and on average, but with a rather small improvement 
of prediction on average, going from 73\% for logit-CPT to $74\%$ for QDT. The same 
issue was encountered by \cite{HMLMurphy} who found that, while their implementation 
of the hierarchical maximum likelihood method improved the reliability of the 
parameter estimates and the log-likelihoods of results at time 2,  the average 
predicted fraction did not improve compared with the one obtained with the 
usual maximum likelihood estimation method. This hints at a hard ``barrier'' 
preventing to improve further the fraction of decisions. Actually, if choices 
are probabilistic, this barrier obtains a natural explanation.

\subsection{Distribution of the predicted fractions}
\label{subsec:distpf}

For a given pair of lotteries $j\in\{1\dots N\}$ and a given decision maker $i$, 
we define the probability $p^i_{A_j}$ with which the lottery $A$ is picked over 
$B$. Likewise, a probability $p^i_{B_j}$ is defined, and $p^i_{B_j}=1-p^i_{A_j}$.

Suppose that the probabilities $p^i_{A_j}$ and $p^i_{B_j}$ are known and stable 
in time. Then the best prediction for the pair of lotteries $j$ is to assume that 
the decision maker will prefer the most likely choice. Consequently, the choice 
regarding lotteries of the pair $j$ can be seen as a Bernoulli trial, with a 
probability of success $p^i_{j}$ larger than 0.5:
\begin{equation}
p^i_{j} = \max \left( p^i_{A_j},p^i_{B_j} \right)
\label{eq:pij}
\end{equation}
 
Let $P^i$ be the fraction of choices predicted correctly for subject $i$. $P^i$ 
corresponds to the fraction of successes in a sequence of $N$ independent Bernoulli 
trials with different probabilities of success. Thus the random variable $P^i$ 
follows a Poisson binomial distribution.

Given the success probabilities $\left(p^i_j\right)_{j\in\{1\dots N\}}$, the 
discrete distribution can be numerically approximated using a discrete Fourier 
transform \cite{poissonPDF} by the following formula:
\begin{equation}
\left\{
\begin{aligned}
&\mathbb{P} \left( P^i=k/N \right) = 
\frac{1}{N+1}\sum_{l=0}^{N}C^{-lk}
\prod_{m=1}^N \left( 1+\left( C^l-1 \right) p^i_j \right) & \;\;\; k \in \left\{ 0\dots N \right\}\\
&C = \exp\left(\frac{2 \omega \pi}{N+1}\right) &
\end{aligned} \right.
\label{wtgwvwqw}
\end{equation}
where $\omega$ stands for the pure imaginary number such that $\omega^2=-1$.

\begin{figure*}[ht]
\centerline{
\includegraphics[width=0.5\textwidth]{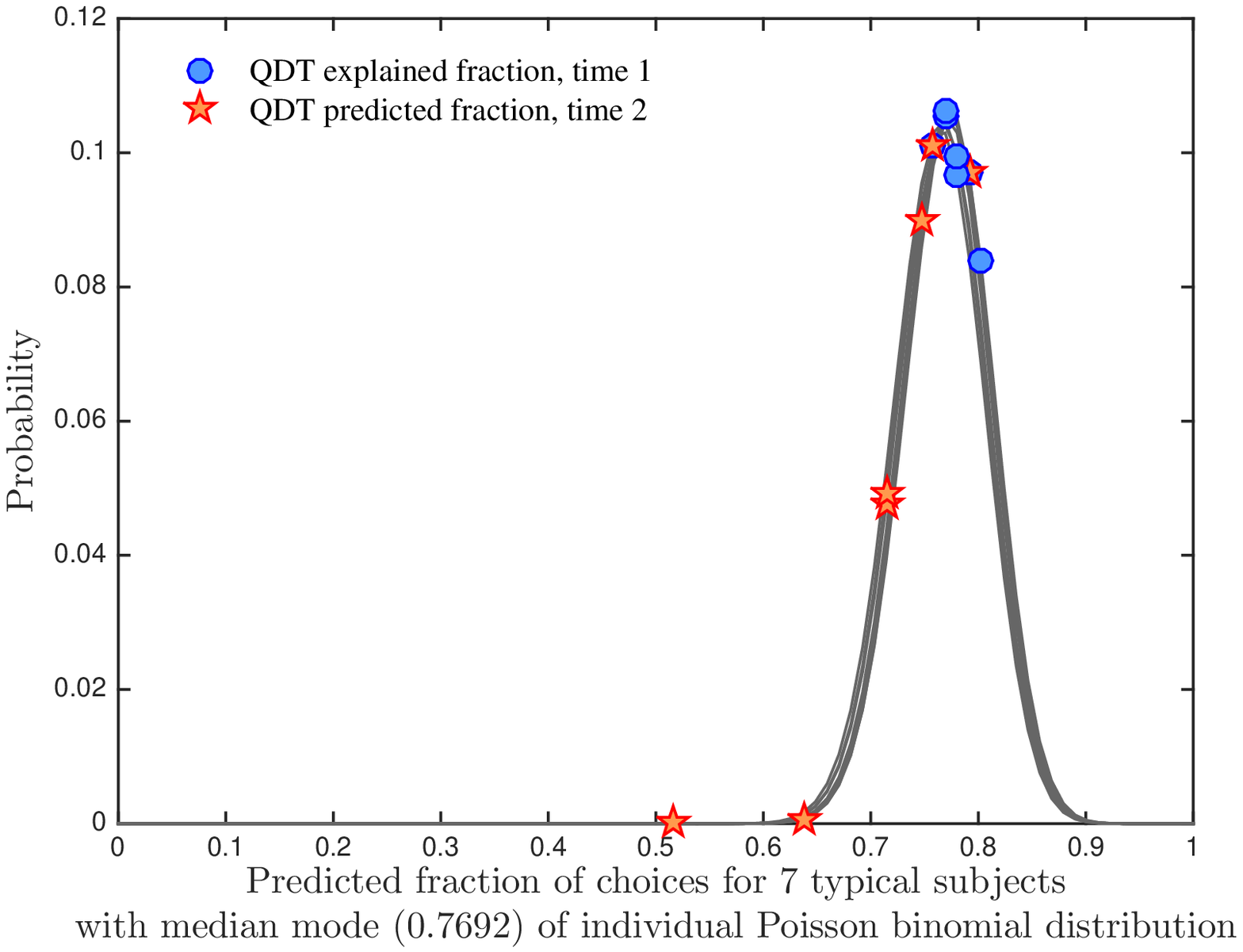}
}
\caption{\small 
Theoretical Poisson binomial distributions of the predicted fraction of choices, 
$P^i$, of a group of 7 typical (with their mode of $P^i$ equal to 0.77, i.e. 
median value within the population: see Figure \ref{fig15}, inset). For these 
theoretical distributions, individual prospect probabilities of the most likely 
choice ($p^i_j>0.5$) for each of the 91 pairs of lotteries $j$'s are estimated 
within the QDT model at time $1$. The observed fractions of choices (i) explained 
at time $1$ i.e. ``in-sample'' (blue circles) and (ii) predicted for time $2$ i.e.
``out-of-sample (red pentagram) with the QDT model, are indicated on the plot. For 
this group of typical decision makers, the theoretical probability to predict more 
than 85\% of the answers is 2.8\%.
}
\label{fig14}
\end{figure*}

\begin{figure*}[ht]
\centerline{
\includegraphics[width=0.6\textwidth]{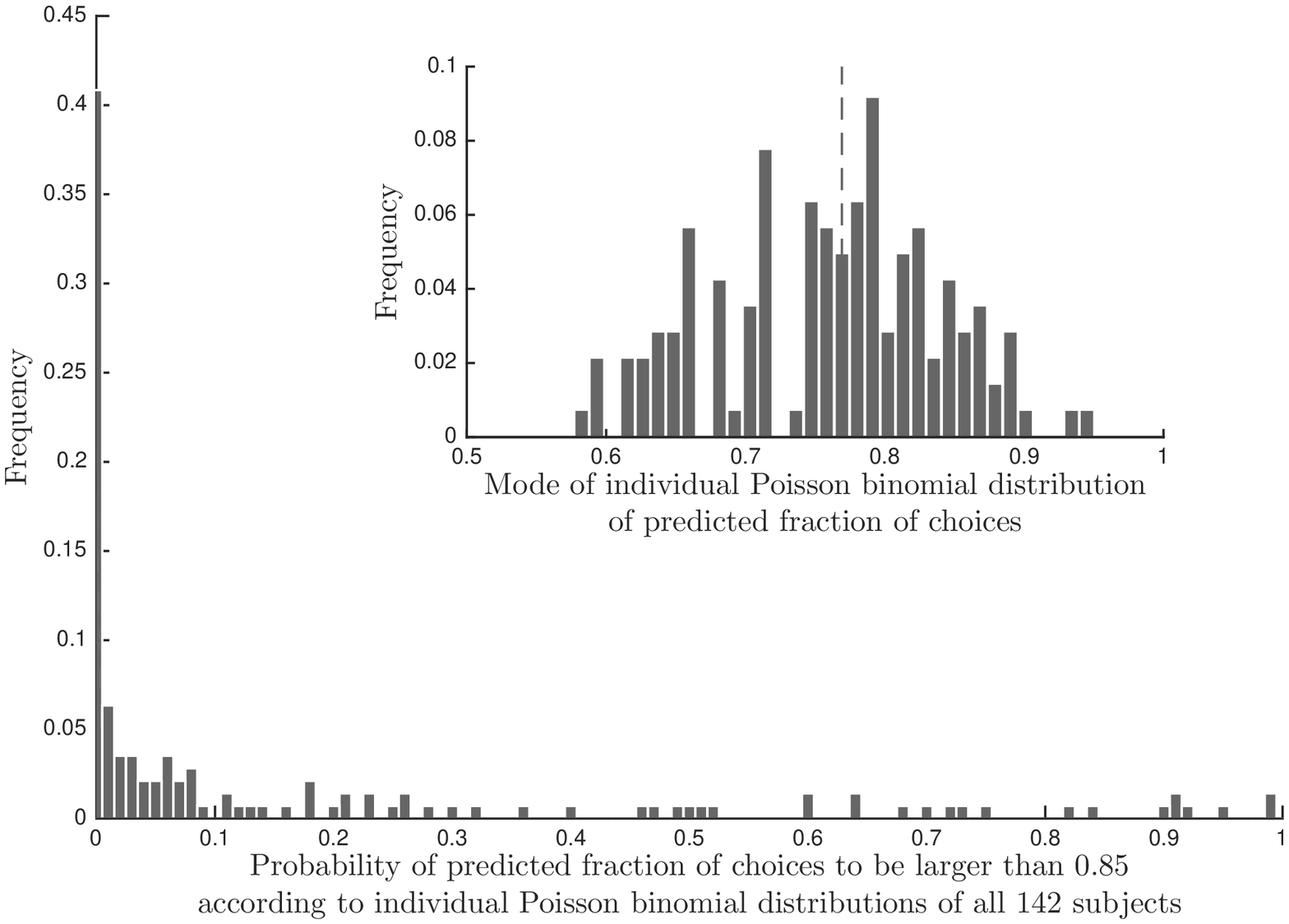}
}
\caption{\small 
{\bf Main plot}: Frequencies, over all 142 subjects, of the probability of the 
theoretical predicted fraction of choices $P^i$ to be larger than 85\%. For 56\% 
of the population (80 subjects), the theoretical  probability to predict more than 
85\% of choices is less than 5\%. {\bf Inset}: Frequencies, over all 142 subjects, 
of the modes of theoretical individual Poisson binomial distributions of the 
predicted fraction of choices, with median value representing a ``typical'' 
decision maker indicated by dashed line.
}
\label{fig15}
\end{figure*}

For the experiment described in Section \ref{subsec:expdesign}, the theoretical 
Poisson binomial distributions of the predicted fraction of choices for a group 
of typical decision makers are plotted in Figure \ref{fig14}. For these distributions, 
individual prospect probabilities of the most likely choice ($p^i_j>0.5$) for each of 
the 91 pairs of lotteries $j$ are estimated with the QDT model at time $1$. These 
values are then inserted in expression (\ref{wtgwvwqw}) to explain (``in-sample) at 
time $1$ and predict (``out-of-sample'') at time $2$ the fraction of correct choices 
(``correct'' in the sense that the choice corresponds to the probability larger than $0.5$ 
as estimated by the QDT calibration). The group of typical decision makers (7 subjects) 
is chosen such that the mode of their theoretical Poisson binomial distribution $P^i$ 
is equal to 0.77, i.e. the median value among the population (see Figure 11, inserted plot). 
For this group of typical decision makers, the theoretical probability to predict more 
than 85\% of the answers is 2.8\%. Similarly to the subjects whose distributions are shown 
in Figure \ref{fig14}, for most decision makers in the experiment, we 
found prospect probabilities for which it was very unlikely to predict more than 
85\% of the answers. Figure \ref{fig15} presents the frequencies, among all 
142 subjects, of the probability of the theoretical predicted fraction of choices 
$P^i$ to be larger than 85\%. From this figure, we can extract the following
representative statistics: for 56\% of the population (80 subjects), the theoretical  
probability to predict correctly more than 85\% of the choices (i.e. $P^i>85\%$) is 
less than 5\%; for 42\% of the decision makers (60 subjects), the probability of 
$P^i>85\%$ is less than 1\%; for 28\% (40 subjects), it is less than 0.1\%. 
Consequently, even if the decision maker's preferences are stable and if the 
estimated probabilities are very accurate, the probabilistic nature of the approach 
does not allow one to improve the choice predictions beyond its theoretical limit 
(which remains randomly distributed).

\subsection{Distribution of predicted fractions at the aggregate level}

Since only one predicted fraction at time $2$ is observed for each subject, it 
is not possible to verify at the individual level whether the predicted fraction 
$P^i$ of choices really follows the Poisson binomial distribution described in 
the previous subsection. However, assuming that the subjects belong to a 
homogeneous population (as discussed in sections  \ref{neytbgww} and \ref{etmyjmk,u5rj}, 
this assumption is not perfect but is useful as a first-order approximation), it 
is possible to approximate the distribution of the predicted fraction throughout 
the population, and to compare it to the histogram of the 142 observed predicted 
fractions at time $2$.

\begin{figure*}[ht]
\hspace{0.75cm}
\centerline{
\begin{subfigure}[h]{.53\textwidth}
\includegraphics[scale=.45]{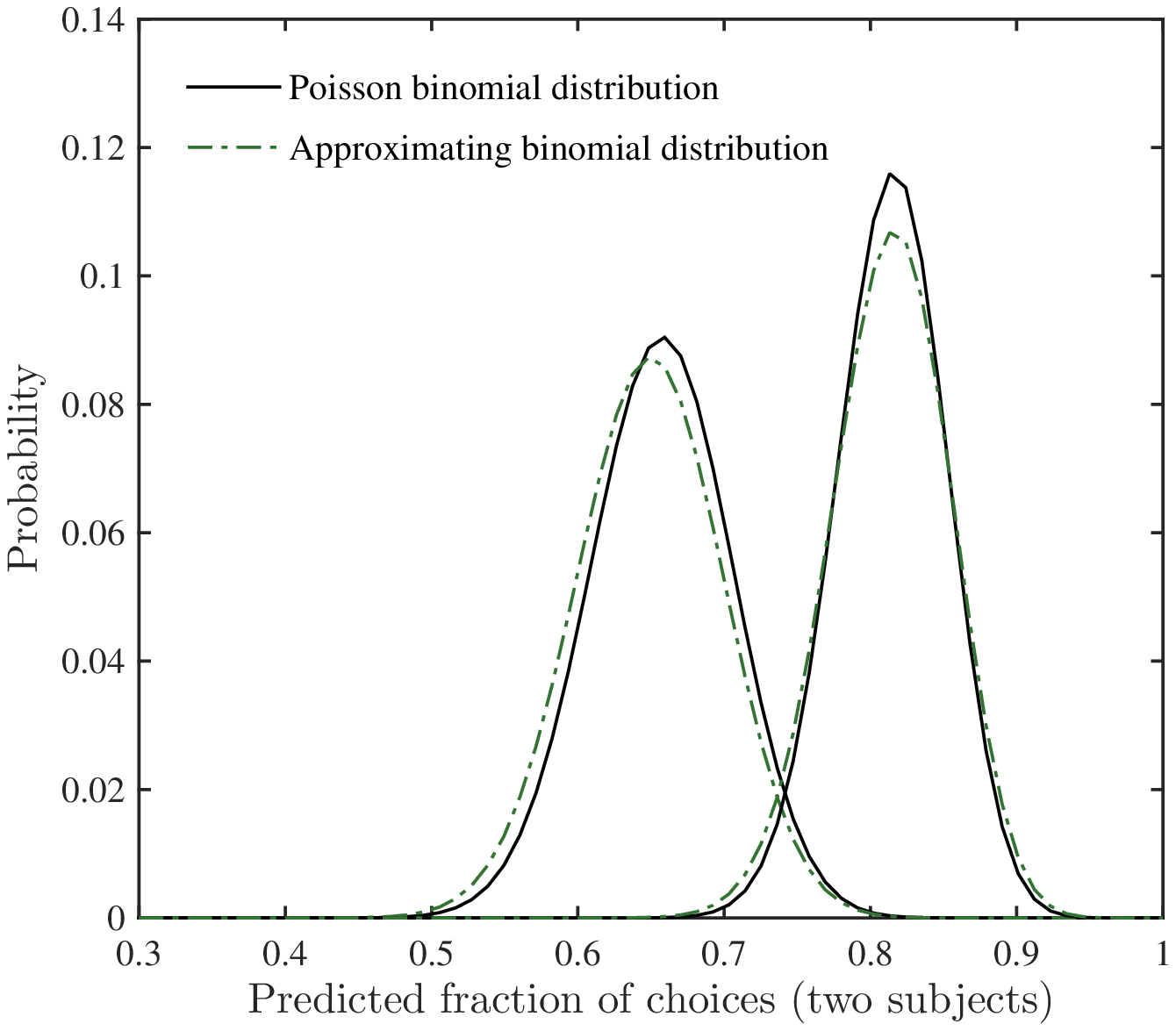} \hspace{1cm}
\end{subfigure}
\begin{subfigure}[h]{.53\textwidth}
\includegraphics[scale=.45]{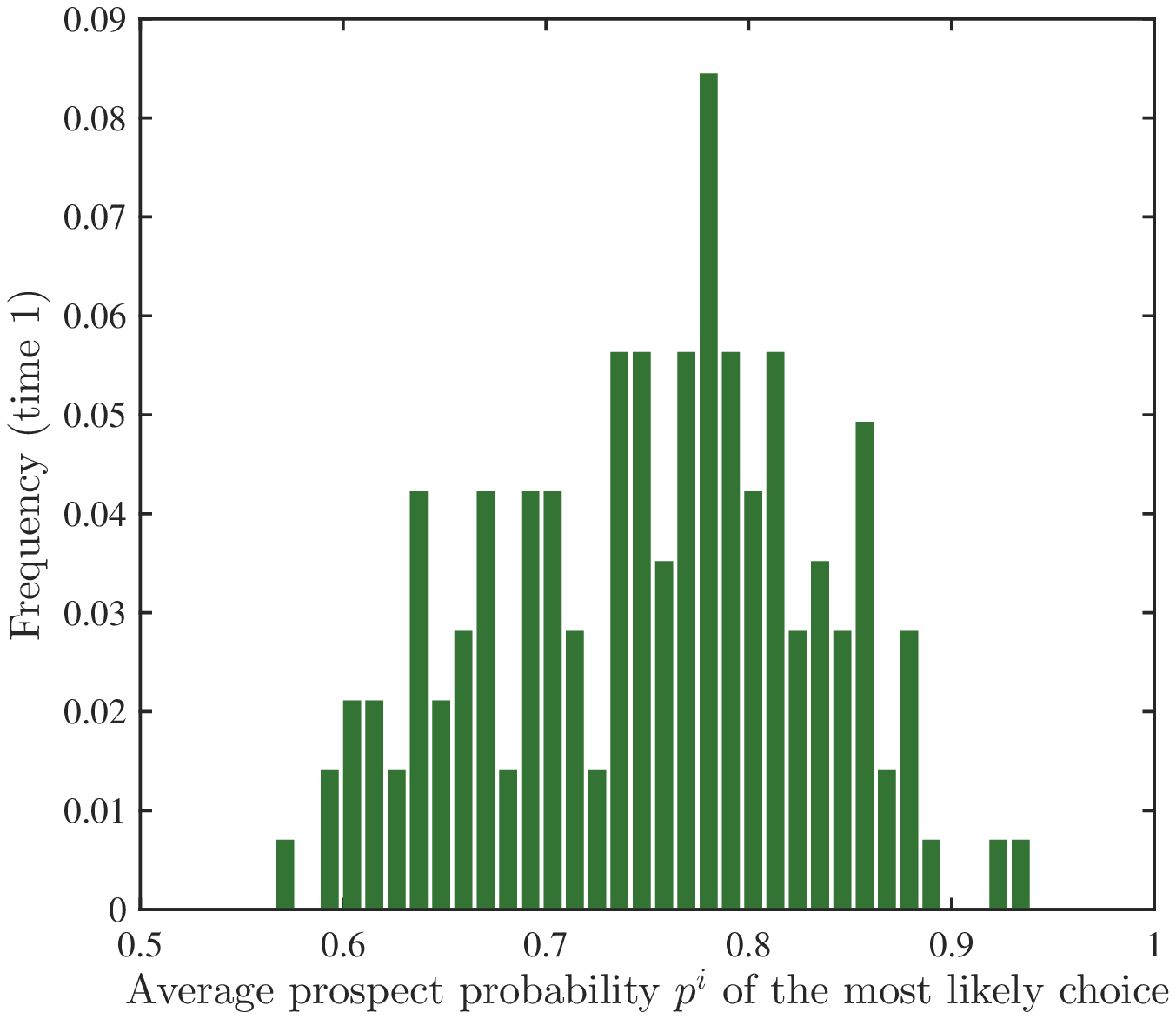}
\end{subfigure}
}
\caption{\small 
{\bf Left}: Theoretical Poisson binomial distributions (black solid line) of 
the predicted fractions of choices for two distinct subjects as described in 
subsection \ref{subsec:distpf}, and their approximating binomial distributions 
(green dash-dotted line).  {\bf Right}: Histogram, over 142 subjects, of their 
observed average prospect probabilities $p^i$ of the most likely choice ($p^i_j>0.5$) 
among 91 pairs of lotteries, estimated at time $1$ with the QDT model (for each 
subject, $p_i$ also corresponds to the mean of her theoretical predicted fraction 
of choices distributed according to the Poisson binomial law, and the approximating 
binomial law).
}
\label{fig16}
\end{figure*}

For this purpose, we now consider that the Poisson binomial distribution of the 
fraction $P^i$ of choices predicted correctly for subject $i$ can be approached 
with the classical binomial distribution $\mathcal{B}\left(p^i,N\right)$, where 
$p^i$ is defined by (see Figure \ref{fig16}, left panel):
\begin{equation}
p^i=
\frac{1}{91}\left\lfloor\sum_{j=1}^{91}p^i_{j}\right\rfloor
\in \left\{ \frac{45}{91},\frac{47}{91}\dots\frac{91}{91} \right\} \; .
\end{equation}

Moreover, we assume that the probability to pick a subject such that 
$P^i\sim\mathcal{B}\left(k/N,N\right)$, with $k\in\left\{45,\dots91\right\}$, is 
equal to the frequency with which $p^i=k/N$ (Figure \ref{fig16}, right panel). 
For each subject, this observed average prospect probability $p^i$ of the most 
likely choice ($p^i_j>0.5$) among 91 pairs of lotteries is estimated at time $1$ 
with the QDT model. These approximations provide accurate representations of the 
results.

\begin{figure*}[ht]
\centerline{
\includegraphics[width=0.5\textwidth]{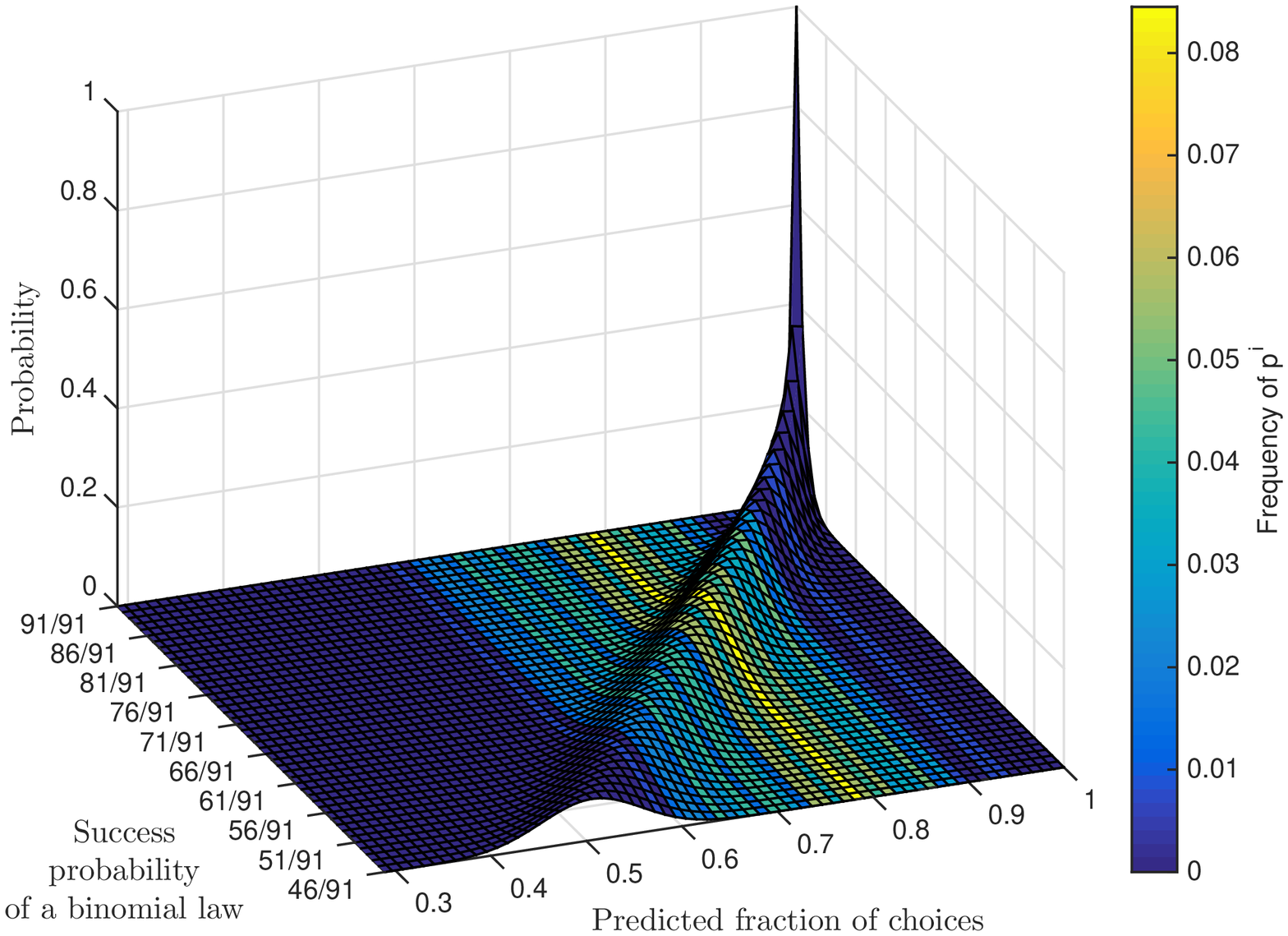}
}
\caption{\small 
Estimation of a theoretical distribution of the predicted fraction of choices throughout 
the population (142 subjects), which is obtained by combining the approximating binomial 
distributions, with success probabilities in the interval $[46/91; 91/91]$, with weights 
determined by the observed frequencies of the average prospect probabilities $p^i$ of 
the most likely choice at time $1$ with QDT model (see Figure \ref{fig18}).
}
\label{fig17}
\end{figure*}

\begin{figure*}[ht]
\centerline{
\includegraphics[width=0.5\textwidth]{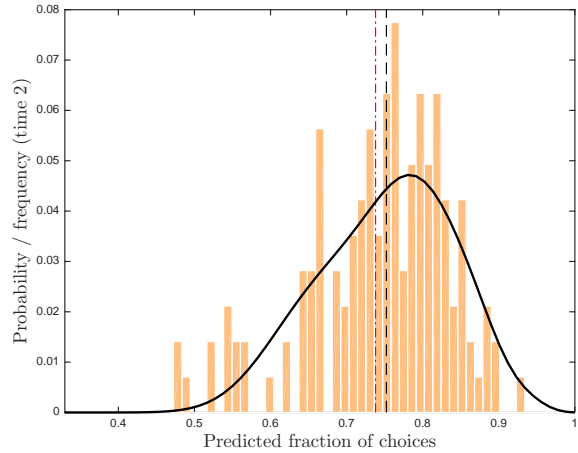}
}
\caption{\small 
Approximated theoretical distribution (black solid line) of the predicted fraction of 
choices throughout the population (142 subjects). The histogram represents the fractions 
of choices correctly predicted ``out-of-sample'' at time $2$ with QDT estimated at time 
$1$. Mean values are indicated by the black dashed line for the theoretical distribution 
and by the red dash-dotted line for the experimental distribution. These values are 
reported in Table \ref{tab:compdist}.
}
\label{fig18}
\end{figure*}

The theoretical distribution of the predicted fraction of choices throughout the 
population (142 subjects) is estimated by approximating binomial distributions, 
with success probabilities in the interval $(0.5; 1]$, which are then weighted by 
the observed frequencies of the average prospect probabilities $p^i$ (see Figure 
\ref{fig17}). Assuming that the prospect probabilities, estimated at time 1, are 
accurate and stable in time and can thus be used at time 2 (to perform an 
``out-of-sample'' prediction), the obtained theoretical distribution of the 
predicted fraction of choices in the population is given by the black solid line 
in Figure \ref{fig18}. The red histogram corresponds to the predicted fractions 
observed at time 2. The approximated theoretical distribution for the predicted 
fraction appears to be close to the experimental one. In particular, both are 
skewed to the left: this suggests that bad predictions at the individual level 
may follow inevitably from the probabilistic nature of the choice.

\begin{table}[ht]
\caption{\small Estimated and experimental moments of the predicted fractions 
throughout the population.}
\label{tab:compdist}
\centering
\begin{tabular}{l c c}
& Approximated distribution & Experimental distribution \\ \hline
Mean & 0.75 & 0.74 \\
Standard deviation & -0.09 & -0.09\\
Skewness & -0.3 & -0.8\\
\end{tabular}
\end{table}

\begin{figure*}[ht]
\centerline{
\includegraphics[width=0.5\textwidth]{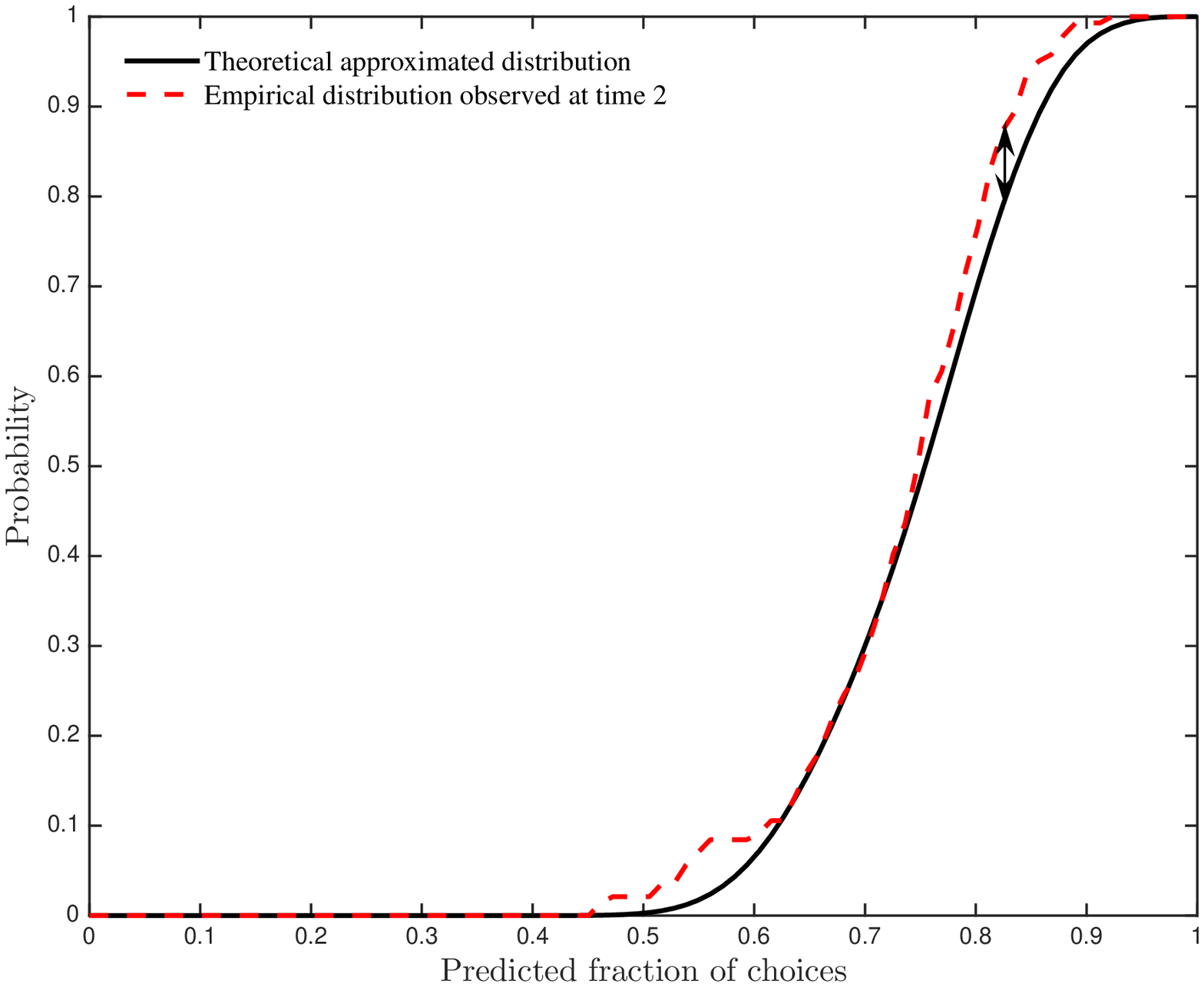}
}
\caption{\small 
Theoretical (black solid line) and experimental (red dashed line) cumulative distribution 
functions (CDF) of the predicted fractions of choices over the population (142 subjects). 
The arrow shows the maximum distance between the two curves (value of the test statistic 
for the Kolmogorov-Smirnov test equal to 0.08).
}
\label{fig19}
\end{figure*}

Performing the Kolmogorov-Smirnov test to compare the theoretical and observed 
distributions of predicted fraction shown in Figure \ref{fig18}, we fail to 
reject at the 5\% significance level the null hypothesis that the experimental 
distribution of the predicted fraction is generated by the theoretical one: the 
p-value is 0.254, and the value of the test statistic is 0.08 (corresponding to 
the maximum distance shown by the arrow in Figure \ref{fig19}.

Table \ref{tab:compdist} and Figure \ref{fig19} compare the estimated cumulative 
distribution function (CDF) of the predicted fraction and the experimental one. Though 
the Kolmogorov-Smirnov test fails to reject the null hypothesis as just mentioned, this 
figure highlights a difference between the two CDF: the theoretical CDF seems to almost 
dominate stochastically the experimental one, i.e., the predicted fractions are less 
good than expected. The reason may be that the model is slightly overfitting. In other 
words, if a subject picks lottery $A$ with probability $p_A>0.5$, and actually chooses 
$A$ at time 1, then maximising the likelihood might lead to overestimating $p_A$, 
thereby overestimating the ``probability of success'' when making the prediction that 
the subject will choose $A$ at time 2.

\section{Further remarks on the link between the probabilistic shift model and 
QDT \label{etmyjmk,u5rj}}

Analysis of the probabilistic shift model in Sections \ref{shifthomog} and 
\ref{neytbgww} was focused on the majority choice. We chose to approximate the 
theoretical probability $p$ in equation (\ref{eq:P_shift_homo}) for homogenous 
population, and $p_1$ and $p_2$ in equations 
(\ref{eq:cnprobs})-(\ref{eq:P_shift_hetero}) with heterogeneity, by an observed 
frequency of the most common choice for a given lottery pair over decision makers. 
Naturally, empirical clustering of the participants of the experiment 
(Figure \ref{fig4}) was based on the same criterion -- their 
tendency to follow/oppose the majority choice. As a consequence, the sign and 
value of the estimated parameters $\al$ and $\bt$, 
equation (\ref{eq:P_shift_hetero_calib}), characterize the two groups only 
with respect to their (dis-)agreement with the majority. In order to make further 
inferences about risk attitudes, they should be estimated within additional models. 
In particular, our QDT model identified higher susceptibility of the ``majoritarian'' 
group towards big losses.

However, it is important to notice that the probabilistic shift model is general. 
Other assumptions to approximate probability $p$ ($p_1$, $p_2$) can be made, e.g. 
with focus on a lottery type, domain of outcomes or a measure of risk. These 
assumptions can guide identifying an alternative clustering of decision makers, 
with different group sizes, and capturing the corresponding risk behaviour.

In this respect, an interesting case arises for a heterogeneous 
model (\ref{eq:P_shift_hetero}) with two groups of equal size: $F=0.5$. Then, 
as Figure \ref{fig5} shows, the best parameters' estimates are 
$\alpha=\beta=1$, leading to a symmetry of the initial model. Moreover, the best 
model can now be expressed as
\begin{equation}
\left\{
 \begin{aligned}
~ p_1 &=2p-p^2\\
~ p_2 &=p^2 ~.
 \end{aligned}  \right.
\label{eq:simplefit}
 \end{equation}
This case of equal-sized groups is illustrated in Figure \ref{fig6} with 
dashed lines. Noteworthy, in a rather large domain of $p$ (from $0.5$ up to not too 
close to $1$), the choice probabilities $p_1$ and $p_2$ of the two groups are such 
that 
 \begin{equation}
\left\{
 \begin{aligned}
~p_1 ~&(p \simeq 0.5) =p+0.25\\
~p_2 ~&(p \simeq 0.5) =p-0.25 ~.
 \end{aligned}  \right.
\label{eq:pmQL}
 \end{equation}

Now recall a fundamental result of QDT, expression (\ref{nhtnbwg}), that the 
probability of a given prospect $\pi_n$ can be decomposed as the sum of two terms, 
the utility factor $f(\pi_n)$ and the attraction factor $q(\pi_n)$. Recall also 
the prediction of QDT called $\it{quarter~law}$ (\ref{eq:QL}) on the magnitude of 
the attraction factor $|q (\pi_n)| \approx 0.25$. Then, the $\pm 0.25$ terms in 
equation (\ref{eq:pmQL}) within a QDT formalism can be interpreted as attraction 
factors, which allows one to account for different risk attitudes among heterogeneous 
decision makers. This would correspond to the non-informative prior for the $q$'s, 
with a positive sign expressing, for example, risk-seeking behaviour or optimism 
of subjects about the value of their choices, and a negative sign expressing risk 
aversion, pessimism or distrust of decision makers.

\section{Conclusion}\label{sec:ccl}

We have analysed an experimental data set comprising 91 choices between two 
lotteries (two ``prospects'') presented in random order made by 142 subjects 
repeated at two separated times. We have proposed an original quantification of 
the choice reversals between the two repetitions, which provides a novel support 
for an intrinsic probabilistic approach to decision making. This has motivated 
us to test for the quantitative performance of a certain parameterisation of 
quantum decision theory (QDT).

As predicted by QDT, we found that variability of individual choices (the average 
rate of choice reversal is $\approx 30\%$, in line with previous studies) is 
accompanied by the stability of the aggregate prospect probabilities (the majority 
choice shifted only for 4 out of 91 lottery pairs). The observed frequency of 
shifts was found in remarkable agreement with the suggested probabilistic model, 
given that it has no adjustable parameters, and the comparison is therefore not a 
fit. Introducing heterogeneity, by differentiating decision makers into two groups, 
we found an excellent quantitative description of the observed frequency of choice 
shifts. For this dataset, the hypothesis of homogeneity is rejected with p-value 
$=1.4\times10^{-7}$, and decision makers are classified as ``majoritarian'' and 
``contrarian'' in proportion $\approx 3:1$.

Presenting a synthetic formulation of the main ingredients of QDT in the Appendix, 
we provided a novel constraint of the attraction factor $q$ for a set of two 
prospects: $|q|\leq\min\left(f,1-f\right)$, where $f$ is the utility factor. The 
new bounds for $q$ are more restrictive than previously considered $[-1; 1]$, and 
are sufficient for insuring the general condition $f+q \in[0,1]$.

This study pioneered a parametric analytical formulation of QDT, integrating 
elements of (a) a stochastic version of Cumulative prospect theory (logit-CPT) 
for the utility factor $f$, and (b) constant absolute risk aversion (CARA) for 
the attraction factor $q$. In essence, this approach allows one to separate risk 
aversion to big losses, and transfer it into the QDT attraction factor. As a 
consequence the loss aversion parameter $\lambda$ was found to be smaller for 
the QDT model, comparing with the benchmark logit-CPT implementation, while the 
values of the other common parameters $(\alpha,\delta,\gamma)$ remained close 
for both models.

Overall, the proposed QDT model improves the results of the logit-CPT model at 
both aggregate and individual levels, and for all considered criteria: explanatory 
power, predictive power, goodness of fit. QDT is especially relevant for the 
``majoritarian'' decision makers ($\approx 75\%$), who show larger susceptibility 
to big losses, as well as higher conviction and consistency of choice. The 
accentuation of the aversion to extreme losses embodied by the QDT attraction 
factor allowed us to noticeably improve the prediction of choices for mixed and 
pure loss lotteries involving big losses. 

At the same time, for most pairs of lotteries, the improvement was rather small. 
This is however hardly unique as there seems to exist a saturation of the average 
predicted fraction of choices at about 73-74\% within the investigated probabilistic 
frameworks. We showed that this hard ``barrier'' is an intrinsic consequence of 
stochasticity in decision making, thus providing additional support for an inherent 
probabilistic component of choice making. 

To quantify the limits of predictability, we proposed the Poisson binomial 
distribution as the theoretical distribution of the individual predicted fraction 
of correct choices. Then, for most decision makers in the experiment, we found 
the prospect probabilities for which it was very unlikely to predict more than 
85\% of the answers. Since only one predicted fraction is observed for each subject 
during the experiment, this theoretical distribution cannot be verified at the 
individual level. Thus, the distribution of the predicted fraction over the whole 
population was approximated with binomial distributions, and was found to be close 
to the experimental distribution of the predicted fractions over the 142 subjects. 
The Kolmogorov-Smirnov test did not reject the null hypothesis that the experimental 
distribution of the predicted fraction is the same as the theoretical one. However, 
the experimental fractions are slightly worse than expected, which may indicate that 
some subjects changed their state of mind, thus being less predictable that we 
assumed. Both distributions are skewed to the left, suggesting an intrinsic 
difficulty in predicting stochastic individual choices. Finally, heterogeneity 
between subjects might also explain these slight discrepancies.

The simplicity of QDT lies in the decomposition (\ref{nhtnbwg}) in which there 
appears the novel quantity, attraction factor. To strengthen the evidence provided 
here, it would be useful to test different forms of the utility factor, as the use 
of the logit-CPT model may be a weakness of the test of QDT, in the sense that we have 
in fact presented a ``joint'' test of two parameterisations:
(i) the use of the logit-CPT model  for the utility factor and (ii) of a constant 
absolute risk aversion (CARA) function for the attraction factor. It is important 
to test other forms for the utility factor, such as rank-dependent utility or regret 
theory which have less parameters, and develop a similar horse-race between these 
models in the original form and with the incorporated attraction factor. Many other 
combinations can be explored.

\section*{Acknowledgements}
We are grateful to R.O. Murphy and R.H.W. ten Brincke for the provided experimental 
data and useful discussions, and to M. Siffert for important remarks. This work was 
partially supported by the Swiss National Foundation, under grant  $105218\_159461$ 
for the project on ``Quantum Decision Theory". One of the authors (V.I.Y.) thanks
E.P. Yukalova for discussions.


\clearpage

\section*{Appendix: Quantum decision theory (QDT)}

Quantum decision theory (QDT)
(Yukalov and Sornette \cite{YSQDT08,YSentropy,YSmathQDT,YSPosOp}) has recently been 
introduced as an alternative formulation to existing decision theories. It 
is based on two essential ideas: (i) an intrinsic probabilistic nature of decision 
making and (ii) a generalization of probabilities using the mathematics of Hilbert 
spaces that naturally accounts for interference and entanglement between choices.

\subsection*{\textbf{Mathematical structure of QDT}}

Let us recall briefly the mathematical construction of quantum decision theory 
(which can be found in more details in \cite{YSmathQDT}).

$\bullet$ Definitions: actions, prospects and state of mind \\

\begin{definition}[Action ring] The action ring 
$\mathcal{A}=\left\{A_n:n=1,2,\dots,N\right\}$ 
is the set of intended actions, endowed with two binary operations: 
\begin{itemize}
\item 
The reversible and associative addition.
\item 
The non-distributive and non-commutative multiplication, which possesses a 
zero element called empty action.
\end{itemize}
\end{definition}

The interpretation of the sum $A+B$ is that $A$ or $B$ is intended to occur. 
The product $AB$ means that $A$ and $B$ will both occur. The zero element is the 
impossible action, so $AB=BA=0$ means that the actions $A$ and $B$ cannot occur 
together: they are disjoint.

\begin{definition}[Composite action and action modes] When an action $A_n$ can 
be represented as a union (i.e. is the sum of several actions), it is referred 
to as composite. Otherwise it is simple.\\
The particular ways $A_{jn}$ of realizing a composite action $A_n$ are called 
the action modes and are disjoint simple elements:
\begin{equation}
A_n=\bigcup_j^{M_n}A_{jn} \;\;\;M_n>1 \; .
\end{equation}
\end{definition}

\begin{definition}[Elementary prospects] An elementary prospect $e_{\alpha}$ 
is an intersection of separate action modes,
\begin{equation}
e_{\alpha}=\bigcap_nA_{{\alpha}n} \; ,
\end{equation}
where the $A_{{\alpha}n}$ are action modes such that $e_{\alpha}e_{\beta}=0$ 
if $\alpha\neq\beta$.
\end{definition}

\begin{definition}[Action prospect] A prospect $\pi_n$ is an intersection of 
intended actions, each of which can be simple (represented by a single action mode) 
or composite
\begin{equation}
\pi_n=\bigcap_j A_{n_j} \; .
\end{equation}
\end{definition}

To each action mode, we associate a mode state $\lgl A_{jn}|$ and its hermitian 
conjugate $\lgl A_{jn}|$. Action modes are assumed to be orthogonal 
and normalized to one, so that $\lgl A_{jn}| A_{kn}\rgl=\delta_{jk}$. This 
allows us to define orthonormal basic states for the elementary prospects:
\begin{equation}
\lgl e_{\alpha}| =
| A_{\alpha 1}\dots A_{\alpha N} \rgl \;\;\;\text{ and }\;\;\;
\lgl e_{\alpha} | e_{\beta} \rgl =
\prod_n\delta_{\alpha_n} \delta_{\beta_n}=\delta_{\alpha\beta} \; .
\end{equation}

\begin{definition}[Mind space and prospect state]\label{def:pstate} The mind space 
is the Hilbert space 
\begin{equation}
\mathcal{M} = {\rm Span}\left\{ | e_{\alpha} \rgl \right\}~.
\end{equation}
For each prospect $\pi_n$, there corresponds a prospect state 
$|\pi_n \rgl \in\mathcal{M}$
\begin{equation} 
|\pi_n \rgl =\sum_\alpha a_{\alpha} | e_{\alpha} \rgl ~.
\end{equation}
\end{definition}

\begin{definition}[Strategic state of mind]\label{def:mstate} The strategic state 
is a normalized fixed state of the mind space $\mathcal{M}$ describing a decision 
maker at a given time:
\begin{equation}
| \psi \rgl = \sum_\alpha c_{\alpha} | e_{\alpha} \rgl ~.
\end{equation}
\end{definition}
The strategic state characterizes a particular decision maker at a given time, 
it includes his/her personal attributes and is related to the information available 
to the decision maker.

\paragraph{$\bullet$ Prospect probabilities}
~~\\
In the context of quantum decision theory, the preferences of a decision maker 
depend on his/her state of mind and on the available prospects. Those preferences 
are expressed through prospect operators. 

\begin{definition}[Prospect operator] For each prospect $\pi_n$, we define the 
prospect operator
\begin{equation}
\hat{P}\left(\pi_n\right) = | \pi_n \rgl \lgl \pi_n | ~.
\end{equation}
\end{definition}

By this definition, the prospect operator is self-adjoint. Its average over the 
state of mind defines the prospect probability $p\left(\pi_n\right)$:
\begin{equation}
p\left(\pi_n\right) = 
\left \lgl \psi\mid\hat{P}\left(\pi_n\right) | \psi\right\rgl ~.
\end{equation} 

The decision maker is more likely to choose the prospect with the highest prospect 
probability. The probabilities should correspond to the frequency with which the 
prospect would be chosen if the choice could be made several times in a same state 
of mind.

By definitions \ref{def:pstate} and \ref{def:mstate}, we can distinguish two terms 
in the expression of $p(\pi_n)$: a utility factor $f(\pi_n)$ and an attraction 
factor $q(\pi_n)$:
\begin{align}
p(\pi_n)& = f(\pi_n) + q(\pi_n) \label{eq:QDTsum1} \; ,\\
f(\pi_n)& = \sum_{\alpha}\mid c^*_{\alpha}a_{\alpha}\mid^2 \label{eq:QDTsum2} \; ,\\
q(\pi_n)& = \sum_{\alpha\neq\beta}c_{\alpha}^*a_{\alpha}a^*_{\beta}c_{\beta} \; .
\label{eq:QDTsum}
\end{align}

Within the framework of quantum decision theory, the utility and attraction terms 
are subjected to additional constraints:
\begin{itemize}
\label{it:constraints}
\item 
$f(\pi_n) \in [0,1]$ and $\sum f(\pi_n)=1$ (normalization of the utility factor)~,
\item 
$q(\pi_n) \in [-1,1]$ and $\sum q(\pi_n)=0$ (alternation property of the quantum factor)~.
\end{itemize}
 
\subsection*{\textbf{Constraint of the attraction factor for a set of two prospects}}
\label{ssubsec:cnQDT}

The QDT formulation for a set of two prospects is now presented, and a constraint 
for the attraction factor $q$ is derived.

In the case of the choice between two lotteries (prospects) $A$ and $B$ (see 
table \ref{tab:gamble}), the constraints on $f$ and $q$ can be written simply:
\begin{equation}
\left\{
\begin{aligned}
p_A&=f_A+q_A\\
p_B&=f_B+q_B\\
q_A&=-q_B\\
f_A&=1-f_B \; .
\end{aligned}
\right.
\end{equation}

The goal being to calibrate quantum decision theory to the decisions made on pairs 
of lotteries, it is important to make some additional assumptions on the prospects 
involved.

Thus, we suppose that the prospects corresponding to the pairs of lotteries presented 
in Table \ref{tab:gamble} can be written as follows:
\begin{equation}
\label{eq:decomp}
\left\{
\begin{aligned}
| A \rgl =a_1 | A1 \rgl +a_2 | A2\rgl \\
| B \rgl =b_1 | B1 \rgl +b_2 | B2 \rgl \; ,
\end{aligned} \right.
\end{equation}
where $|A1\rgl$, $|A2\rgl$, $|B1\rgl$ and $|B2\rgl$ are orthogonal action mode states 
(this decomposition might be linked to the coexistence of belief and disbelief as 
suggested in \cite{YSPosOp}, but the precise content of these action mode states will 
not be specified here).

We write the state of mind as
\begin{equation}
| \psi \rgl  = 
c_{A_1} | A1 \rgl + c_{A_2} |A2 \rgl + c_{B_1} | B1 \rgl  + c_{B_2} | B2 \rgl
\end{equation}
and we denote by $f_{A_1}$ and $f_{A_2}$ the following quantities
\begin{equation}
\label{eq:F}
f_{A_1}= | c_{A_1}^*a_1 |^2;\;\;\;f_{A_2}= | c_{A_2}^*a_2 |^2~.
\end{equation}
Then, the utility factor $f_A$ satisfies
\begin{equation}
\label{eq:F2}
f_A = f_{A_1} + f_{A_2}~.
\end{equation}
Moreover, according to equation (\ref{eq:QDTsum}), the attraction factor is such 
that
\begin{equation}
q_A = c_{A_1}^*a_1a_2^*c_{A_2} + c_{A_2}^*a_2a_1^*c_{A_1} =
2\re{\left( c_{A_1}^*a_1a_2^*c_{A_2} \right) } \; .
\end{equation} 
Consequently, we can introduce the {\it uncertainty angle} $\Dlt^A$ \cite{YSentropy} 
such that
\begin{equation}
\label{eq:unangle}
\begin{split}
q_A = 2\sqrt{f_{A_1}f_{A_2}} \cos\left(\Delta^A\right) \; .
\end{split}
\end{equation}
Moreover, equations (\ref{eq:F}) and (\ref{eq:F2}) imply that there exists some 
$x\in[0,1]$ such that
\begin{equation}
f_{A_1} = xf_A\text{ and }f_{A_2} = \left(1-x\right)f_A
\end{equation}
So, for some $x\in[0,1]$, we have that
\begin{equation}
q_A = 2f_A\sqrt{x\left(1-x\right)} \cos\left(\Delta^A\right) \; .
\end{equation}
In particular, $|q_A|\leq f_A$, and the same reasoning for the lottery $B$ gives 
$|q_B|\leq f_B=1-f_A$.
Consequently, given that $|q_A|=|q_B|$, we obtain that 
\begin{equation}
\label{eq:adconstr}
|q_A| \leq \min \left(f_A,1-f_A \right) \; .
\end{equation}
Therefore, assumption (\ref{eq:decomp}) leads to a constraint for the attraction 
factor of quantum decision theory for a set of two prospects, which is 
given by equation (\ref{eq:adconstr}). The bounds for $q_A$ are found to be more 
restrictive than previously considered $[-1,1]$, and are sufficient to insure the 
general condition $f_A+q_A\in[0,1]$.

\subsection*{\textbf{Analytical formulation for the calibration of QDT}}

Under the assumptions done in the previous subsection, the formulation of QDT for 
choices between two lotteries $A$ and $B$ should be such that:
\begin{itemize}
\item $f_A=1-f_B$ (normalization)
\item $q_A=-q_B$ (alternation)
\item $q_A=\min\left(f_A,f_B\right)\cos\left(\Delta^A\right)$ (uncertainty factor)
\end{itemize}
The two next subsections address the parametrisation chosen for the utility term 
$f$ and the attraction term $q$.

\subsubsection*{\textit{Utility term and stochastic cumulative prospect theory}}

Since the f-factor should represent a normalized utility, it is a natural choice 
to make it correspond to a stochastic version of cumulative prospect theory (CPT). 
Prospect theory was introduced by \cite{KaTversky} and is now the most famous 
alternative to expected utility theory. Within this framework, the outcomes are 
transformed through a value function $v$, and the probabilities are modified 
through a non-additive weighting function $w$. Moreover the value function separates 
gains and losses, where the notions of gains and losses are defined with respect to 
a reference point, here assumed to be zero. Cumulative prospect theory (CPT) is a 
variation of prospect theory, in which the weighted probabilities for outcomes of 
same sign should sum up to $1$ \cite{KaTversky2}.

With this CPT model, for the simple pairs of lotteries shown in Table \ref{tab:gamble}, 
a lottery $A$ is valued by
\begin{equation}
\tilde U_A = \left\{
\begin{aligned}
w\left(p^A_1\right)v\left(V^A_1\right)&+\left(1-w\left(p^A_1\right)\right)v\left(V^A_2\right)
&\text{ if $sign(V^A_1)=sign(V^A_2)$}\\ 
w\left(p^A_1\right)v\left(V^A_1\right)&+w\left(p^A_2\right)v\left(V^A_2\right)
&\text{ if $sign(V^A_1)\neq sign(V^A_2)$}\\ 
\end{aligned}
\right.
\label{eq:utiPT}
\end{equation}
where $V^A_1$, $V^A_2$ have been ordered such that:
\begin{itemize}
\item $V^A_1\geq V^A_2$ if both are positive.
\item $V^A_1\leq V^A_2$ if both are negative.
\end{itemize}
The value function $v$ is chosen to be convex in the domain of losses and concave 
in the domain of gains. These properties reflect commonly observed behavioural 
patterns: risk aversion concerning gains, and risk seeking behaviour with respect 
to losses.

For probability weighting, different formulations tend to suggest an inverse-S 
shaped function, so that small probabilities are overweighted and large probabilities 
underweighted. 

In the present article, the value function is a power function with the same 
exponent $\alpha$ in the gain and the loss domains with a kink at $0$ quantified 
by the loss aversion coefficient $\lambda$:
\begin{equation}
\label{eq:prospv}
v\left(x\right) = \left\{ \begin{aligned}
x^\alpha\;\;\;\;\;\;x\geq0\;\;\;\alpha>0
\\-\lambda\left(-x\right)^\alpha\;\;\;\;\;x<0\;\;\;\lambda>0 \; .
\end{aligned}
\right.
\end{equation}

For probability weighting, a function known as the Prelec II weighting function 
was chosen \cite{Prelec}. It includes two parameters: $\dlt$ controls the general 
elevation of the curve, and $\gamma$ controls its curvature,
\begin{equation}
\label{eq:prospw}
w(p) = \exp\left(-\delta\left(-\ln\left(p\right)\right)^\gamma\right)~,
\;\;\;\;\;\;\delta>0\;\;\;\gamma>0~.
\end{equation}

Given the individual preferences and characteristics of a decision maker, CPT aims 
at predicting the choice of a decision maker, assuming it to be deterministic in 
the sense that the decision corresponds to the option that maximises the outcome 
values weighted by subjective probabilities. In order to account for the ubiquitous 
observation of choice stochasticity, CPT can be combined with a probabilistic choice 
function. In particular, the stochastic version of CPT incorporates the probabilistic 
deviation of a decision maker from the option that maximises the choice criterion 
with respect to alternative options. Stott \cite{Menagerie} investigated several 
possible combinations and confirmed that a power value function (\ref{eq:prospv}) 
combined with a Prelec II weighting function (equation \ref{eq:prospw}) is a good 
choice. Several formulations of stochastic CPT were ranked in \cite{Menagerie} and 
the logit-power-Prelec II combination appeared to offer a good tradeoff between 
quality of fit and number of parameters, when it is supplemented by a logit choice 
function (referred to as logit-CPT). The probability predicted by stochastic CPT of 
picking an option $A$ over $B$ is assumed to coincide with the f-factor of QDT and 
is given by
\begin{equation}
\label{eq:ffac} 
f_A = \frac{1}{1+e^{\varphi\left(\tilde U_B-\tilde U_A\right)}}~,
\end{equation}
where $\varphi$ is a steepness parameter.

According to this formulation of the $f$-factor given by stochastic cumulative 
prospect theory, the utility factor of QDT can be characterized by five parameters: 
two for the value function ($\alpha$, $\lambda$), two for the weighting function 
($\gamma$, $\delta$) and one for the choice function ($\varphi$). This formulation 
of the value and probability weighting functions as well as the stochastic component 
is identical to that used by \cite{HMLMurphy} and allows for a straightforward 
comparison of their results with our calibration of QDT. Indeed, when the attraction 
factor $q$ is vanishing, QDT then reduces to stochastic CPT.

\subsubsection*{\textit{Attraction factor}}

As for the attraction factor, we have
\begin{equation}
\label{eq:genqfac}
q_A = \min\left(f_A,f_B\right)\cos\left(\Delta^A\right) \; ,~~~~~ q_A+q_B=0~.
\end{equation}

The main issue is then to find a good parametrisation of the uncertainty factor 
$\cos\left(\Delta^A\right)$ that adds useful information, without adding too many 
parameters. In the current study, we replaced the cosine by 
\begin{equation}
\label{eq:cosine}
\cos\left(\Delta^A\right) \longrightarrow \tanh\left(a\left(U_A-U_B\right)\right)~,
\end{equation}
where $U_A$  and $U_B$ are utilities associated with the lotteries $A$ and $B$ that 
need to be specified, and $a$ is either an additional parameter or a pre-defined 
constant. Thus,
\begin{equation}
\label{eq:qfac}
q_A = \min\left(f_A,f_B\right)\tanh\left(a\left(U_A-U_B\right)\right) \; .
\end{equation}
This formulation satisfies automatically the alternation condition $q_A=-q_B$. To be 
specific, we assume that $U$ is the constant absolute risk aversion (CARA) function 
for an initial wealth of 100 corresponding to the amount given to the subjects at the 
beginning of the experiment:
\begin{equation}
\label{eq:CARA100}
U\left(V\right) = 1 - e^{-\eta\left( 100+V \right) } \; .
\end{equation}
With this formulation, as Figure \ref{fig8} illustrates, $q_A$ tends to be 
negative when the lottery $A$ involves big losses and is compared to a lottery $B$ 
with more moderate losses.

\clearpage

\bibliographystyle{plainnat}

\end{document}